\documentclass[sigconf,twocolumn, screen, 10pt]{acmart}

\usepackage{booktabs} 
\usepackage[english]{babel}
\usepackage{xcolor}
\usepackage{xspace}
\usepackage{colortbl}
\usepackage{pifont}
\usepackage{subfigure}
\usepackage{graphicx}
\usepackage[inline]{enumitem}
\usepackage{booktabs}
\usepackage{threeparttable}

\usepackage{multirow}
\usepackage{fancybox} 
\usepackage{amsmath}
\usepackage{hhline}
\usepackage{amsthm}
\usepackage{comment}
\usepackage{tabularx}
\usepackage[utf8]{inputenc}
\usepackage{bbm}
\usepackage{amsfonts}
\usepackage{pythonhighlight}
\usepackage{fancyhdr}
\usepackage{soul}

\renewcommand\footnotetextcopyrightpermission[1]{} 
\setcopyright{none}

\usepackage{hyperref}
\hypersetup{
  colorlinks=true,      
  linkcolor=blue,       
  citecolor=magenta,    
  filecolor=cyan,       
  urlcolor=red          
}

\definecolor{boxcolor}{gray}{0.9}

\usepackage[linesnumbered,ruled,noend,scleft]{algorithm2e}
\usepackage{float}
\usepackage{algorithmicx}

\usepackage{xcolor}
\definecolor{cardinal}{rgb}{0.77, 0.12, 0.23}
\definecolor{fuchsia}{RGB}{218,112,214}

\SetCommentSty{mycommfont}
\newenvironment{denseitemize}{
\begin{itemize}[topsep=2pt, partopsep=0pt, leftmargin=1.5em]
  \setlength{\itemsep}{2pt}
  \setlength{\parskip}{0pt}
  \setlength{\parsep}{0pt}
}{\end{itemize}}

\definecolor{codegreen}{rgb}{0,0.6,0}
\definecolor{codegray}{rgb}{0.5,0.5,0.5}
\definecolor{codepurple}{rgb}{0.58,0,0.82}
\definecolor{backcolour}{rgb}{0.95,0.95,0.92}

\lstdefinestyle{codestyle}{
    commentstyle=\color{purple}\slshape,
    keywordstyle=\color{black}\bfseries,
    numberstyle=\color{codegray},
    basicstyle=\ttfamily\mdseries\scriptsize,
    emph={AdaEmbed,emb_agent},
    emphstyle={\color{black}\bfseries},
    breakatwhitespace=false, 
    frame=lines,      
    rulecolor=\color{codegray},  
    breaklines=true,                 
    captionpos=b,                    
    keepspaces=true,                 
    numbers=left,                    
    numbersep=5pt,                  
    showspaces=false,                
    showstringspaces=false,
    showtabs=false,                  
    tabsize=2
}
\definecolor{eclipseStrings}{RGB}{42,0.0,255}
\definecolor{eclipseKeywords}{RGB}{127,0,85}
\colorlet{numb}{magenta!60!black}

\lstdefinelanguage{json}{
    basicstyle=\scriptsize\ttfamily,
    commentstyle=\color{eclipseStrings}, 
    stringstyle=\color{eclipseKeywords}, 
    emph={AdaEmbed,emb_agent},
    emphstyle={\color{black}\bfseries},
    numbers=left,
    numberstyle=\scriptsize,
    stepnumber=1,
    numbersep=8pt,
    showstringspaces=false,
    breaklines=true,
    frame=lines,
    string=[s]{"}{"},
    comment=[l]{:\ "},
    morecomment=[l]{:"},
    literate=
        *{0}{{{\color{numb}0}}}{1}
         {1}{{{\color{numb}1}}}{1}
         {2}{{{\color{numb}2}}}{1}
         {3}{{{\color{numb}3}}}{1}
         {4}{{{\color{numb}4}}}{1}
         {5}{{{\color{numb}5}}}{1}
         {6}{{{\color{numb}6}}}{1}
         {7}{{{\color{numb}7}}}{1}
         {8}{{{\color{numb}8}}}{1}
         {9}{{{\color{numb}9}}}{1}
}

\def\ie{{i.e.}}
\def\eg{{e.g.}}

\def\name{EMA\xspace}

\definecolor{darkgreen}{RGB}{0, 200, 0}

\usepackage{tikz}

\newcommand*\blackcircled[1]{\tikz[baseline=(char.base)]{
            \node[shape=circle,fill,inner sep=1pt] (char) {\textcolor{white}{#1}};}}
\algnewcommand{\LeftComment}[1]{\Statex \(\triangleright\) #1}

\renewcommand{\paragraph}[1]{%
  \vspace{2mm} \noindent \textbf{#1} \hspace{0.5mm}
}

\AtBeginDocument{%
  }

\definecolor{darkgreen}{RGB}{0, 200, 0}

\usepackage{tikz}

\begin{document}
\setcounter{page}{1}

\renewcommand{\headrulewidth}{0pt}
\renewcommand{\footrulewidth}{0pt}

\pagestyle{fancy}
\fancyhf{}
\fancyfoot[C]{\thepage}

\title{\huge{\name: Efficient Model Adaptation for Learning-based Systems}}

\author{Daiyang Yu$^*$}
\thanks{$^*$ Indicates equal contribution. Work done while at UIUC}
\affiliation{
  \institution{University of Illinois Urbana-Champaign}
  \country{USA}
}

\author{Xinyu Chen$^*$}
\affiliation{
  \institution{University of Illinois Urbana-Champaign}
  \country{USA}
}

\author{Yihan Zhang$^*$}
\affiliation{
  \institution{University of Illinois Urbana-Champaign}
  \country{USA}
}

\author{Yan Liang}
\affiliation{
  \institution{The Hong Kong University of Science and Technology}
  \country{Hong Kong, China}
}

\author{Yaqi Qiao}
\affiliation{
  \institution{University of Illinois Urbana-Champaign}
  \country{USA}
}

\author{Fan Lai}
\affiliation{
  \institution{University of Illinois Urbana-Champaign}
  \country{USA}
}

\begin{abstract}
Machine learning (ML) is increasingly applied to optimize system performance in tasks such as resource management and network simulation. Unlike traditional ML tasks (e.g., image classification), networked systems often operate in heterogeneous, long-running, and dynamic environment states, where input conditions (e.g., network loads) and operational objectives can shift over time and across settings. Existing learning-based systems offer little support for adaptation, resulting in costly model training, extensive data collection, degraded system performance, and slow responsiveness.

This paper presents EMA, the first model adaptation system supporting learning-based systems to adapt to evolving environments with minimal operational overhead. EMA takes a system-driven, data-centric approach that accommodates diverse system and model designs while addressing two key deployment challenges. First, it reduces expensive model training by introducing state transformers that align the input state of a new environment with previously similar states, allowing models to warm-start adaptation. Second, it addresses the often-overlooked yet costly process of data labeling---collecting ground truth for exploring and training on various system decisions---by prioritizing labeling high-utility data while balancing the tradeoff between training and labeling cost. Evaluations on eight representative learning-based systems show that EMA reduces adaptation costs (e.g., GPU training time) by 14.9--42.4\% while improving system performance (e.g., network throughput) by 6.9--31.3\%.

\end{abstract}

\maketitle

\section{Introduction}
\label{sec:intro}

Machine learning (ML) has been an increasingly powerful tool for optimizing systems and networks, with applications spanning traffic engineering~\cite{dote-nsdi23, xu2023teal} and video streaming~\cite{pensieve-sigcomm17, puffer-nsdi20} over wide-area networks (WANs), flow scheduling~\cite{li2024-QCLIMB, dhukic2019-flux, jury-eurosys25}, network simulation~\cite{mimicnet-sigcomm21, yang2022deepqueuenet}, and resource management in the cloud~\cite{firm-osdi20, flash-mlsys24, decima-sigcomm19, sinan-asplos21}. 
These systems employ a broad spectrum of learning techniques, from classical models such as linear regression (LR)~\cite{sinan-asplos21} and random forests~\cite{li2024-QCLIMB}, to deep learning (DL) approaches~\cite{dote-nsdi23} including reinforcement learning (RL)~\cite{pensieve-sigcomm17} and even large language models (LLMs)~\cite{netllm-sigcomm24}. By learning latent correlations from operational data, learning-based systems have demonstrated superior automation and decision quality over traditional hand-crafted heuristics.

Despite their success, the effectiveness of these systems critically depends on how well their models align with the underlying \emph{environment state}---the joint distribution of system inputs, workloads, and objectives. Unlike traditional ML tasks (\eg, image classification), system deployments span diverse and evolving environments. Variations may arise from differences in infrastructure (e.g., cluster sizes and hardware types~\cite{puffer-nsdi20}), supporting workloads (e.g., flow and job size distributions~\cite{jury-eurosys25}), and operational objectives (e.g., service-level
objectives across tenants~\cite{firm-osdi20}). 
Even within the same environment, its states can change over time (\eg, traffic dynamics, workload evolution,  and machine upgrades).  Without timely model adaptation, system performance can degrade sharply, sometimes by over 80\% (\S\ref{sec:background}), a fundamental challenge echoed in Microsoft's operational experience~\cite{autosys-atc20}.

However, existing learning-based systems have largely overlooked the design for their efficient (model) adaptation. Supporting each new environment often requires tuning on extensive system traces, which can take many hours~\cite{fluxion-nsd23, autothrottle-nsdi24} and cost thousands of dollars in GPU resources~\cite{netllm-sigcomm24}.
As a result, adaptation is slow, expensive, and often impractical for production systems that demand rapid response to runtime dynamics (\S\ref{sec:background}). 
Recent advances have explored plugin-based systems support~\cite{flash-mlsys24}, but most are ad-hoc and task-specific, such as normalizing network signals by bandwidth capacity~\cite{jury-eurosys25}. Often, they require intrusive system modifications (e.g., inserting meta-learners into model architectures~\cite{flash-mlsys24}).

Worse still, real-world system adaptation often involves costly data labeling, a process that collects ground-truth feedback (\ie, label) for the model to explore the impact of various system decisions and converge (\eg, alternative network attack remediation strategies). These labels are highly environment-specific, often requiring replaying workloads in controlled deployments, building and running simulators, or soliciting expert annotation (\eg, from network operators~\cite{caravan-osdi24}). This has been largely overlooked in prior work yet can dominate end-to-end adaptation overhead (\S\ref{sec:background}).

This paper introduces \name, a model adaptation system that takes a data-centric approach to automate the recurring, daunting task of adapting learning-based systems to diverse, evolving environments, with only a few lines of integration code (\S\ref{sec:overview}). \name leverages the insight that systems are typically long-running, accumulating a repository of trained models and operational data across deployments. When a new adaptation request arises, such as deploying a learned system in a new environment state, \name identifies a prior environment with similar states and repurposes its model and data as \emph{operational knowledge} to warm-start adaptation (e.g., training). During adaptation, it monitors system performance and selectively acquires new labels.

\name addresses two fundamental challenges in practical deployments. First, it must enable efficient and generalizable transfer of operational knowledge across environments to avoid ``reinventing the wheel'', despite variations in tasks, model architectures, and telemetry. Unlike existing adaptation techniques that often require intrusive system changes~\cite{flash-mlsys24}, \name applies a lightweight, one-shot transformation to system input data entirely outside the model and system logic. It projects inputs into a latent state space, identifies a prior environment with similar state distributions, and derives a transformation to align the new input distribution to the source. Bridging this state discrepancy enables \name to reuse trained model weights and data  (\S\ref{sec:transformer}).

Second, achieving efficient adaptation requires tackling the inherent cost tension between model training and data labeling. 
Labeling a large volume of data inflates collection costs but can reduce training costs by exposing the model to broader data coverage, enabling faster convergence and better accuracy. Moreover, labeling costs vary across inputs (e.g., evaluating scheduling policies when allocating 2$\times$ versus 10$\times$ more machines to a job), and the helpfulness of labeled data to improve system performance shifts as training progresses.  
\name introduces a cost-aware labeling agent that prioritizes labeling data expected to yield larger model performance improvement per unit cost (\S\ref{sec:labeler}). At runtime, \name orchestrates training and labeling through a cost-benefit lens, determining when and how much to label to maximize overall cost-effectiveness. It further manages caching of model and state repository for future reuse across requests (\S\ref{sec:orchestrator}).

We evaluated \name on seven representative systems from SIGCOMM, NSDI, and OSDI:   Flux~\cite{dhukic2019-flux} for flow size prediction using LR models; DOTE~\cite{dote-nsdi23} for WAN traffic engineering using DL; MimicNet~\cite{mimicnet-sigcomm21} for datacenter flow simulation using LSTMs; FIRM~\cite{firm-osdi20} for microservices resource management with RL; Pensieve~\cite{pensieve-sigcomm17} for adaptive bitrate streaming (ABR) with RL; and NetLLM~\cite{netllm-sigcomm24} for ABR and cluster job scheduling with LLMs. Our evaluations (\S\ref{sec:eval}) show that, compared to state-of-the-art efforts~\cite{flash-mlsys24, caravan-osdi24}, \name cuts adaptation costs (\eg, GPU time) by 14.9--42.4\% and accelerates system adaptation by 2.3--15.3$\times$, while improving post-adaptation system performance, such as network throughput and user experience in video streaming, by 6.9--31.3\%. 

In summary, we make the following contributions:
\begin{denseitemize}
    \item We present the first general model adaptation system supporting diverse learning-based systems; 

    \item We introduce a novel data-centric approach to repurpose operational knowledge while optimizing data efficiency;

    \item We evaluate \name across seven learned systems, showing its substantial gains to augment practical deployments.
\end{denseitemize}

\paragraph{Ethics:} This work does not raise any ethical issues.

\section{Background and Motivation}
\label{sec:background}

Building a performant learning-based system is inherently iterative and labor-intensive~\cite{autosys-atc20}. Developers must curate large volumes of system inputs (e.g., job demands or network traces) and collect corresponding system feedback (e.g., task completion times under various schedules) as training labels, via replay runs in controlled deployments,  simulations, or soliciting expert annotations~\cite{caravan-osdi24}. They then navigate a broad design space of models (e.g., LR, LSTMs, or LLMs) to balance system performance and runtime overhead.  

Even after careful design and tuning, operating conditions rarely remain stable.  The input state of a learning-based system often varies across three key environmental dimensions: \emph{infrastructure} (e.g., network capacity or cluster size), \emph{workloads} (e.g., flow size or job arrival distributions), and \emph{operational objectives} (e.g., latency-throughput-fairness tradeoffs). Even within the same deployment, these states can shift over time, such as due to resource and workload changes~\cite{puffer-nsdi20}.

\begin{figure}[t]
  \centering
  {
    \subfigure[Sizeless~\cite{eismann2021sizeless} performance varies.\label{fig:sizeless-over-different-env}]{\includegraphics[width=0.36\linewidth]{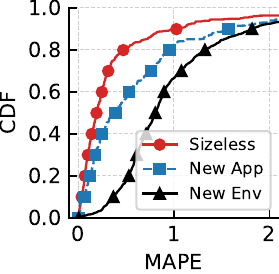}}
    \hfill
    \subfigure[FIRM~\cite{firm-osdi20} performance varies across applications. \label{fig:firm-scatter}]{\includegraphics[width=0.58\linewidth]{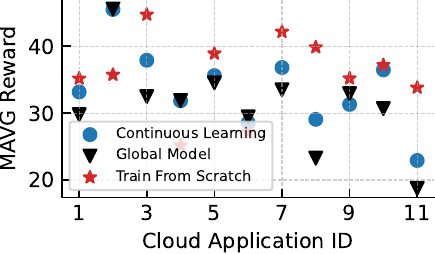}}
  }
  \vspace{-.4cm}
  \caption{Deployments of learning-based systems often face varying deployment environments and operational objectives, demanding efficient adaptation.}
  \label{fig:motivation:fig2}
\end{figure}

\paragraph{Environmental variations lead to performance drift.}
To achieve high performance, learning-based systems typically tune their ML models on environment-specific traces. This, however, comes with a double-edged cost: environmental variations can easily render large system performance degradation. To illustrate, we analyze Sizeless~\cite{eismann2021sizeless}, a DL-based system for predicting the execution time of serverless functions. Using real CloudBandit~\cite{CloudBandit} production workloads, we evaluate its performance across three representative scenarios: (i) deployment in the same environment as training, (ii) supporting new serverless functions, and (iii) supporting the same function on new infrastructure (e.g., hardware). Figure~\ref{fig:sizeless-over-different-env} shows the system performs well in its original environment, but its decision quality drops largely on new functions (\ie, ``New App'') and collapses further under new infrastructure (\ie, ``New Env'').

We observe similar trends in FIRM~\cite{firm-osdi20}, a learning-based microservice resource scheduler (Figure~\ref{fig:firm-scatter}): deploying a global model without application-specific adaptation degrades system performance by 4.5--79.2\% across applications.

\begin{figure}[t]
  \centering
  \includegraphics[width=0.99\linewidth]{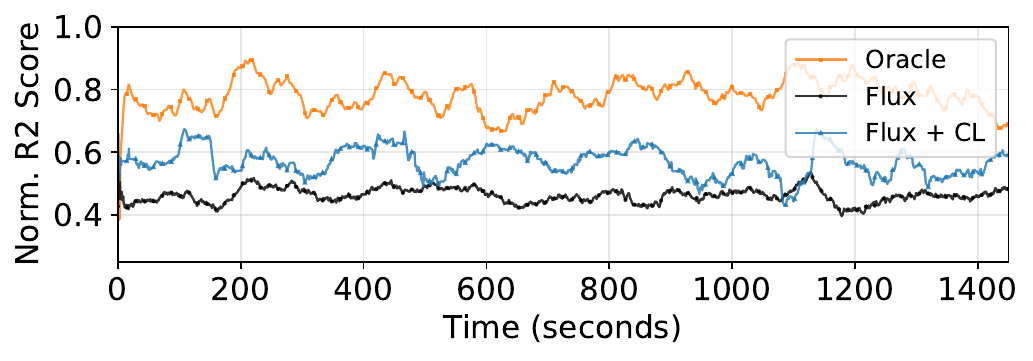}
  \vspace{-.4cm}
  \caption{In FLUX (flow size prediction for network scheduling), system performance varies as job arrivals (flow characteristics). Even with continuous learning (CL), the model performs far below the oracle setting.\protect\footnotemark[1]} 
  \label{fig:flux-online_no_ema}
\end{figure}

\paragraph{System adaptation introduces efficiency and quality challenges.}
Mitigating performance drift requires model adaptation that is fast and cost-effective, as adaptation latency directly translates to degraded system turnaround, SLO violations, and revenue loss (\S\ref{eval:e2e}). 
Recent systems, such as Caravan~\cite{caravan-osdi24}, explore online learning for in-network models. However, even model fine-tuning, especially with the growing adoption of DL- or LLM-based AIOps~\cite{confucius-sigcomm25}, can take hours~\cite{netllm-sigcomm24}. In latency-sensitive settings such as real-time network monitoring and attack response~\cite{leo-nsdi24}, adaptation must complete in minutes or even seconds (e.g., Figure~\ref{fig:flux-online_no_ema} and more in Section~\ref{eval:e2e})\footnote{The oracle represents a hypothetical system model trained with access to all future data. Detailed experiment settings are available in Section~\ref{exp:setup}.}, creating a fundamental mismatch between tuning cost and operational timescales.

Moreover, practical deployments extend beyond the single-environment assumption of online learning and require efficient cross-environment adaptation. For example, due to the monolithic control logic (e.g., 1,034 service knobs), adapting cloud microservice management systems to new services can take 24 hours, during which system performance degrades by up to 26\%~\cite{fluxion-nsd23}. At scale, where thousands of services undergo frequent software updates, hardware changes, and workload shifts, these adaptation costs accumulate rapidly.

Even beyond efficiency concerns, adaptation poses fundamental challenges to system performance. As shown in Figure~\ref{fig:firm-scatter}, continuously fine-tuning the FIRM model for individual applications can improve performance in some cases, but is not consistently reliable: the resulting models can underperform those trained from scratch, and even the offline global model. This behavior reflects \emph{negative transfer}~\cite{long2015learning}, where operational knowledge learned from one environment misaligns with the state distribution of the target environment. We observe similar phenomena in FLUX (Figure~\ref{fig:flux-online_no_ema}), and across many learning-based systems (\S\ref{eval:e2e}).

\begin{figure}[t]
  \centering
  {
    \subfigure[Data labeling and training cost. \label{fig:data-percentage}]{\includegraphics[width=0.48\linewidth]{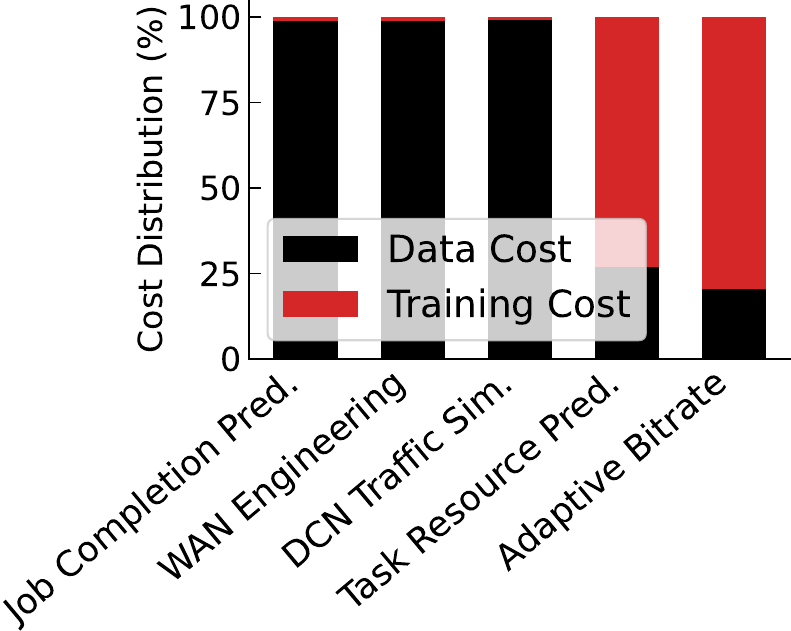}}
    \hfill
    \subfigure[Norm. data labeling cost.\label{fig:data-cdf}]{\includegraphics[width=0.48\linewidth]{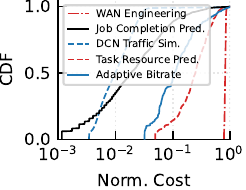}}
  }
  \vspace{-.4cm}
  \caption{Data labeling is costly in learning-based systems and can vary across samples (\ie, system inputs).}
  \label{fig:collecct-data}
\end{figure}

\begin{figure*}[t] 
  \centering
  \includegraphics[width=1.0\textwidth]{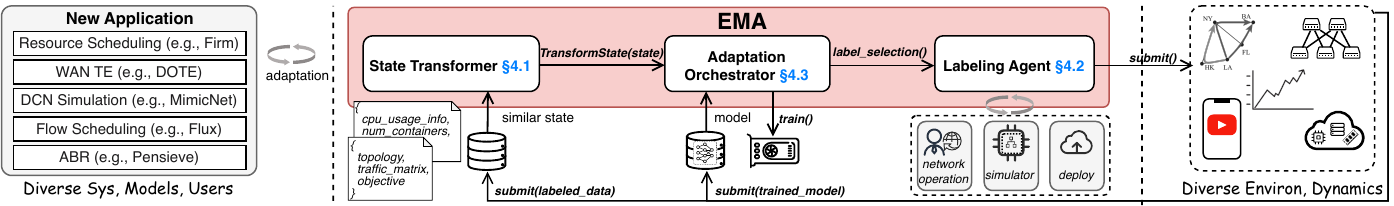} 
  \caption{\name repurposes operational knowledge of similar environments to optimize systems adaptation.} 
  \label{fig:sys-arc}
\end{figure*}

\paragraph{Data labeling introduces new efficiency bottlenecks.} 
Beyond training overhead, data labeling emerges as another barrier to efficient adaptation. Learning-based systems are tightly coupled to environment-specific factors: while logging system input (e.g., job arrivals or resource demands) is cheap, obtaining corresponding performance feedback (labels) is not. Labels require executing system decisions to observe outcomes, which is often time-consuming (e.g., via simulation), resource-intensive (e.g., via controlled deployments), and sometimes infeasible due to access constraints.  

To quantify this cost, we study five representative learning-based systems: DOTE~\cite{dote-nsdi23} (WAN traffic engineering), MimicNet~\cite{mimicnet-sigcomm21} (network simulation), Flash~\cite{flash-mlsys24} and Sinan~\cite{sinan-asplos21} (QoS-aware cloud resource management), and NetLLM~\cite{netllm-sigcomm24} (adaptive bitrate streaming and cluster job scheduling). As shown in Figure~\ref{fig:data-percentage}, data labeling can account for a substantial fraction, and in some cases the majority, of end-to-end adaptation cost. We quantify costs using equivalent Google Cloud Platform instance hours. For example, labeling 20,000 data points for adapting Sinan's model consumed over 6 hours executing on VMs~\cite{autothrottle-nsdi24}. The imbalance is more pronounced for large models. Fine-tuning NetLLM for cluster job scheduling, even with low-rank adaptation and on a small TPC-H workload, requires roughly 340 A100 GPU hours (about \$458), while data labeling---running jobs under different resource configurations to observe execution time---costs approximately 572 VM hours (about \$476). Importantly, these costs are not one-time investments: they recur when the environment changes, such as under new workload distributions, cluster reconfigurations, or shifts in operational objectives (e.g., efficiency-fairness tradeoffs).

Moreover, data labeling costs are highly heterogeneous across input samples (Figure~\ref{fig:data-cdf}), with some being orders of magnitude more expensive due to differences in simulation complexity or resource demands. For instance, evaluating the impact of allocating $10\times$ additional resources to a job incurs substantially higher cost than exploring a $2\times$ allocation. 

\paragraph{Limitations of existing advances for model adaptation.}
Adaptation mechanisms in today's learning-based systems remain fragmented and narrowly scoped. First, most approaches are tightly coupled to specific tasks or model classes (\S\ref{sec:related}), relying on ad hoc techniques such as manual systems feature normalization~\cite{jury-eurosys25}, intrusive model modifications (e.g., inserting meta-learners)~\cite{flash-mlsys24, netllm-sigcomm24}, or continuous learning strategies confined to a single deployment or environment~\cite{caravan-osdi24}. 
Second, existing efforts largely overlook the cost of data labeling and its interaction with model training. In practice, labeling is frequently the dominant and most heterogeneous component of adaptation cost, yet current systems lack mechanisms to reason about when, what, and how much data to label under operational constraints.

As a result, developers are forced to repeatedly engineer bespoke adaptation pipelines for each new system, workload, or environment. This not only increases engineering complexity and prolongs system turnaround time, but also often leads to suboptimal post-adaptation performance (\S\ref{sec:eval}). 
\section{\name Overview}
\label{sec:overview}

This paper presents \name, the first model adaptation system that enables efficient adaptation for diverse learning-based systems and model designs with minimal operational overhead. Our study of representative networked systems (\S\ref{sec:background} and \S\ref{sec:eval}) identifies three key requirements for practical adaptation: 
(i)~\emph{Computation-efficient}: minimizing training time to preserve system responsiveness and control resource cost (e.g., GPU hours) for scaling to many environments;
(ii)~\emph{Data-efficient}: reducing expensive data labeling while maintaining or improving model quality; and
(iii)~\emph{Generalizable and deployable}: supporting a wide range of systems and models with minimal system changes and human intervention.

\name achieves these goals through a \emph{system-driven, data-centric approach} that operates entirely outside the model or system logic, acting directly on system input state data as a complementary layer between learning-based systems and their underlying runtime (e.g., resource orchestrators for RL-based schedulers~\cite{firm-osdi20}). Appendix~\ref{app:apis} details \name's interfaces, demonstrating that existing systems can be integrated with only a few lines of code changes.

\paragraph{System Components.}
As shown in Figure~\ref{fig:sys-arc}, 
\name processes model adaptation requests online, such as those issued by performance monitoring systems like Caravan~\cite{caravan-osdi24}, whether for deployment in a new environment, serving new users, or addressing performance degradation. Upon receiving an adaptation request:
\blackcircled{1} The \emph{State Transformer} locates a source environment with a similar state distribution and applies a transformation (e.g., via a distribution shift matrix) to bridge the new state's input to the prior.
\blackcircled{2} The \emph{Adaptation Orchestrator} initializes the target model by transferring the learned model weights of the source environment and forwarding transformed inputs to the tuning backend (\eg, PyTorch), warm-starting both model and data.
\blackcircled{3} During training, the orchestrator monitors progress and invokes the \emph{Labeling Agent} to select inputs for labeling (\eg, via interactions to online labeling systems like Caravan~\cite{caravan-osdi24}), balancing training efficiency against labeling cost.
\blackcircled{4} After training, the updated model and the metadata of request's state (\eg, data distributions) are stored in the \emph{StateStore}, enabling caching, reuse, and faster adaptation for future requests.

\section{\name Design}
\label{sec:design}

We now describe the design of \name, a data-centric adaptation system that jointly optimizes adaptation efficiency and quality for learning-based systems. Rather than treating adaptation as a monolithic procedure, \name decomposes it into a coordinated pipeline with three phases:
(1)~pre-training state adaptation to reduce subsequent training computation (\S\ref{sec:transformer}); 
(2)~in-training adaptation to optimize the data efficiency in labeling (\S\ref{sec:labeler}); and 
(3)~cross-round orchestration to balance training (computation) and data costs while managing adaptation states for future reuse at scale (\S\ref{sec:orchestrator}). 
Together, these phases address the system-level tension between \emph{quality} (high post-adaptation model accuracy) and \emph{efficiency} (low cost and latency), which prior adaptation systems~\cite{flash-mlsys24, caravan-osdi24} have largely treated in isolation.

\subsection{State Transformer: Bridge State Gaps}
\label{sec:transformer}

Adapting ML models to diverse and evolving system states is essential for sustaining system performance. 
Relying on a global model degrades accuracy under state variations, while naive continuous learning or fine-tuning is resource-intensive, slow to react to runtime changes, prone to instability, and confined to a single online deployment (\S\ref{sec:background}).

Our key insight is that learning-based systems are typically long-running and deployed across correlated environments. Over time, they naturally accumulate a repository of trained models and observed environment states, forming a reusable body of \emph{operational knowledge}: a memory of how the system behaves under diverse conditions. Repurposing this knowledge provides a principled warm start for new environments, with the potential to enable faster adaptation and higher post-adaptation quality (e.g., by leveraging broader coverage from previously encountered, similar states).

However, reusing such knowledge introduces three key deployment challenges. 
First, although conceptually related to transfer learning (TL)~\cite{long2015learning}, existing TL techniques are largely designed for DL models. Yet, practical learning-based systems may rely on lightweight, non-DL models, where adaptation efficiency is still critical for real-time operations,  cost savings, and scaling out (\S\ref{eval:e2e}). 
Second, TL methods often require intrusive changes to system pipelines or model architectures, or incur high runtime overhead, such as inserting auxiliary meta-learners~\cite{flash-mlsys24}. 
Finally, effective reuse hinges on selecting an appropriate source environment, but the diversity of features, workloads, and system tasks makes source selection highly challenging. Poor source selection can induce negative transfer, slowing convergence or even degrading performance (\S\ref{sec:background}).

\begin{figure}[t]
  \centering
  \includegraphics[width=0.99\linewidth]{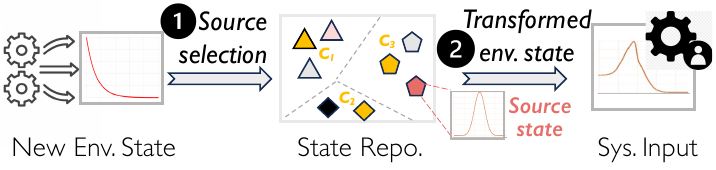}
  \caption{\name identifies a source state from prior deployments, and then transforms the new environment state to closely align with it, enabling the previously learned system model to be effectively repurposed.} 
  \label{fig:tca-workflow}
\end{figure}

\paragraph{Regime-aware State Transformation.}
To address these challenges, \name introduces a lightweight \emph{state transformer} that operates entirely at the system input boundary, without modifying models or systems.
As shown in Figure~\ref{fig:tca-workflow}, \name identifies a promising source environment via state matching, then transforms the target environment's input state to align with the source before feeding it to the model.

\name enables state transformation by aligning the input state distributions of a source environment $\mathbb{S}$ and a target environment $\mathbb{T}$ in a shared latent space (e.g., a Gaussian kernel space $\mathbb{G}$).
Specifically, \name obtains a transformation matrix $W$ that maps raw system inputs into $\mathbb{G}$ such that the transformed source and target state distributions are closely aligned:
$
\min_{W} \; \mathrm{MMD}\big( \Phi(\mathbb{S}), W \,\Phi(\mathbb{T}) \big),
$
where $\Phi(\cdot)$ denotes the kernel-induced feature mapping into $\mathbb{G}$, and $\mathrm{MMD}(\cdot,\cdot)$ measures the distributional discrepancy between two sets of states. Obtaining $W$ involves matrix transformation operations like transfer component analysis~\cite{Pan2011} and we add details in Appendix~\ref{sec:privacy}. 
Our transformation allows a model trained under the source environment to interpret target inputs as if they originated from a familiar operating regime.
As such, 
\name loads the learned model weights of the source as the starting point for subsequent in-training adaptation on the transformed target states (\S\ref{sec:labeler}).

However, learning-based systems introduce another unique challenge: system states often follow \emph{highly skewed and regime-diverse} distributions. For example, a small fraction of large jobs or elephant flows dominate resource consumption and system performance.
A single global transformation risks collapsing these rare but performance-critical states into dominant regimes, obscuring precisely the behaviors that matter most for adaptation.

To address this, \name employs a \emph{regime-aware state transformation} strategy.
After projecting source and target states into the latent space $\mathbb{G}$, \name partitions each environment's states into \emph{operational regimes} via clustering, capturing distinct system behaviors (e.g., short vs. long flows, latency-sensitive vs. throughput-oriented jobs).
\name then performs group-wise alignment: for each target regime, it identifies the most similar source regime with smallest $MMD$ and derives a regime-specific transformation. By aligning corresponding regions of the state space independently, \name preserves rare but influential behaviors while avoiding distortion from dominant regimes. 

\begin{figure}[t]
  \centering  
    \subfigure[Adaptation Efficiency. \label{fig:speedup-pear}]{
      \includegraphics[width=0.47\linewidth]{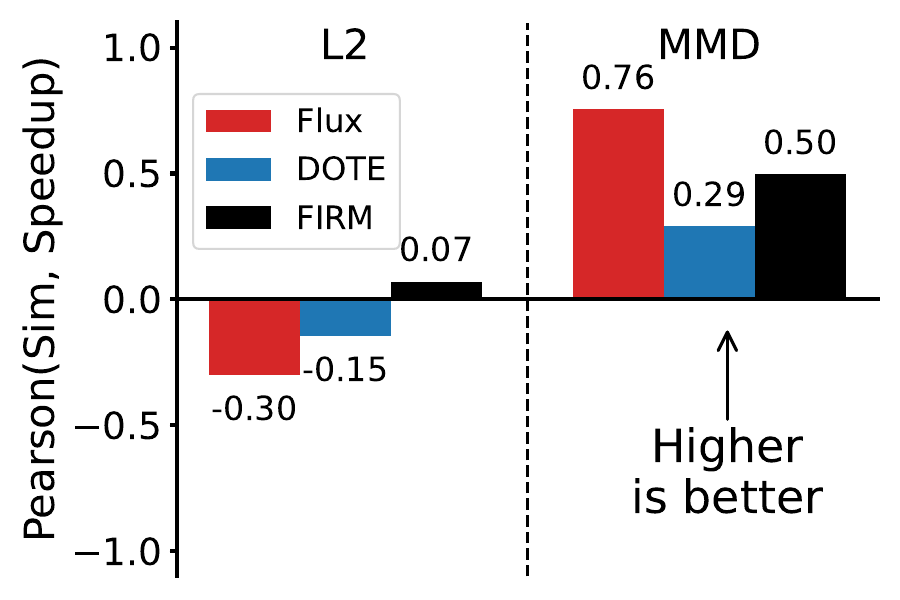}
   }
   \hfill
   \subfigure[Post-adaptation Model Accuracy. \label{fig:sim-pear}]{
      \includegraphics[width=0.47\linewidth]{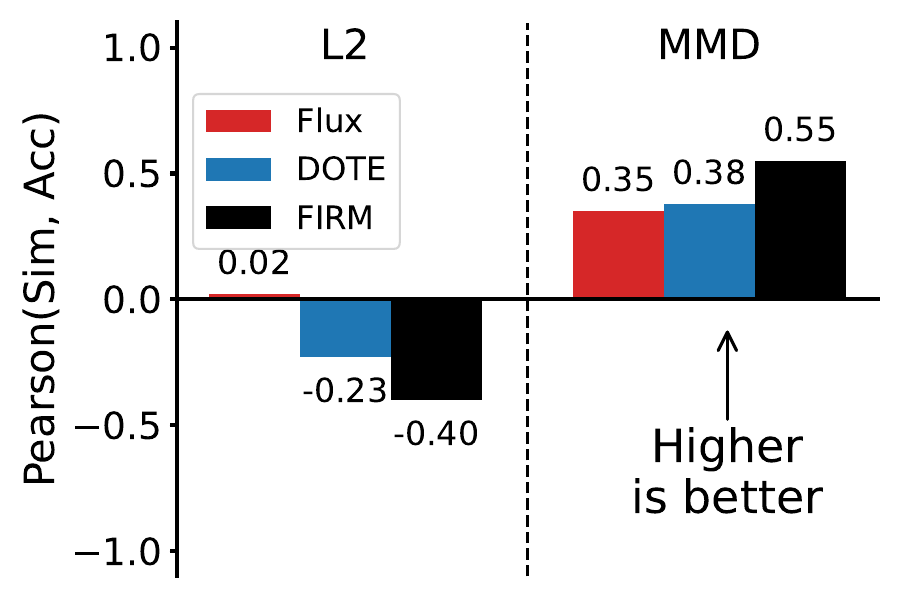}
   }
   \vspace{-.4cm}
  \caption{MMD similarity exhibits a stronger positive Pearson correlation with improvements on adaptation efficiency and post-adaptation system performance, compared to feature-agnostic L2 distance. }
  \label{fig:mmd-improvment}
\end{figure}

Our state transformer yields three key system advantages.
(1) \emph{Model independence}: since alignment occurs solely at the input data layer, it requires no system changes and additional information.
(2) \emph{Broad applicability}: the same mechanism applies to diverse state representations, including numeric telemetry, time series, graphs, and embeddings, while supporting a wide range of models used in systems, from LR and LSTM to RL and LLMs (\S\ref{eval:e2e}).
(3) \emph{Principled source selection}: $MMD$ provides a task-agnostic, distributional similarity metric for selecting source environments.
As shown in Figure~\ref{fig:mmd-improvment}, smaller $MMD$ strongly correlates with faster convergence and higher post-adaptation accuracy, while large discrepancies often lead to negative transfer. 
Accordingly, \name selects the source environment with the minimum $MMD$, outperforming traditional distance metrics such as L2-distance.

\begin{figure}[t]
  \centering  
  \subfigure[Overhead of DOTE. \label{fig:overhead-Dote}]{
      \includegraphics[width=0.47\linewidth]{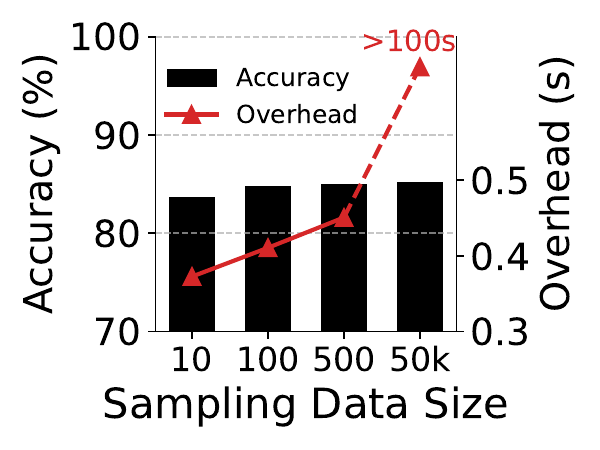}
  }
  \hfill
  \subfigure[State transformer is efficient. \label{fig:transform-scalability}]{
      \includegraphics[width=0.47\linewidth]{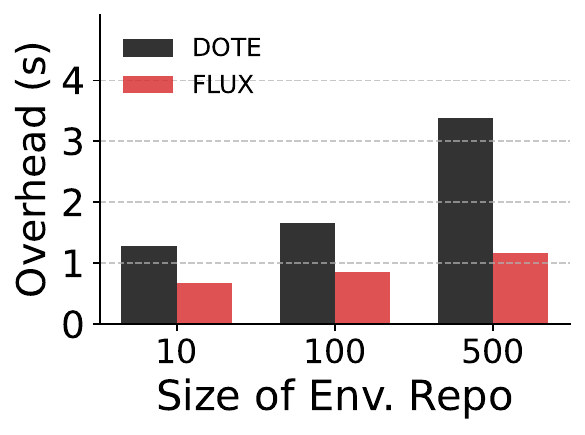}
  }
  \vspace{-.4cm}
  \caption{\name reduces transformation overhead through data sampling (a), and scales efficiently to hundreds of environment states in the repository (b).}
  \label{fig:sample-tca}
\end{figure}

\paragraph{Scaling State Transformation.}
While lightweight at runtime, state transformation incurs a one-time preprocessing overhead: (i) computing the transformation matrix $W$ that maps the target environment into the source environment, and (ii) selecting the most suitable source environment and the subsequent regime-aware transformation require calculating and comparing pairwise MMD distances. 
Both operations involve matrix computations whose cost grows with the volume of state data and the number of candidate environments. As shown in Figure~\ref{fig:sample-tca}, naively computing MMD over large traces can take minutes, and this overhead scales with the number of source candidates. To scale, \name reduces (1) the per-comparison cost by \emph{sampled state transformation}, and (2) the number of comparisons by \emph{clustered matching}.

First, as shown in Figure~\ref{fig:sample-tca}, using excessive data points to compute the transformation matrix and $MMD$ significantly increases preprocessing cost while providing diminishing accuracy gains. 
Leveraging this insight, \name performs sampled state transformation, computing $MMD$ and the transformation matrix by sampling only a small subset of source and target states. Specifically, let $F_m(x)$ denote the empirical distribution estimated from $m$ samples and $F_N(x)$ the true distribution over the full dataset of size $N$. The concentration theorem~\cite{10.14778/2733004.2733022} bounds their maximum deviation $D=\sup_x |F_m(x)-F_N(x)|$.\footnote{To guarantee $D\le\epsilon$ with confidence $\delta$ (default 95\%), it suffices to use
$
m \ge -\frac{1}{2\epsilon^2}\ln\left(\frac{1-\delta}{2}\right).
$
}
This yields only a few thousand samples, even for populations with millions, significantly cutting preprocessing cost while preserving fidelity.

Second, to avoid enumerating all target-source candidate comparisons, \name clusters candidate environments offline into groups using K-medoids~\cite{NIPS2017_a8345c3b} and incrementally updates the clustering online (\S\ref{sec:orchestrator}). 
Unlike the widely used K-means, which requires embedding heterogeneous system features, our 
K-medoids design directly leverages the pairwise MMD distance to choose real environments as medoids (Figure~\ref{fig:tca-workflow}), ensuring interpretability and generalizability. 
However, clustering into too few clusters creates large groups, increasing the number of intra-cluster comparisons, while too many clusters inflate the cost of identifying the closest medoid. Assuming a total of $M$ environment candidates, each request requires $K$ operations to identify the nearest cluster (i.e., matching to $K$ medoids), and then on average $M/K$ comparisons within the chosen cluster. The total expected matching cost is thus $C(K) = K + M/K$. Minimizing $C(K)$ yields the optimal number of clusters to use: $K = \sqrt{M}$.

As shown in Figure~\ref{fig:transform-scalability}, these optimizations enable low-latency transformation (< 4 seconds) even under hundreds of environment states in the repository. We further show that \name can achieve adaptation improvements with as few as a dozen candidate states in the repository (\S\ref{eval:ablation}).

\subsection{Labeling Agent: Optimize for Data} 
\label{sec:labeler} 

While state transformation provides a strong warm start that accelerates adaptation (\S\ref{eval:e2e}), learning-based systems often require additional model tuning to capture environment-specific behaviors. In this in-training adaptation phase, the cost of labeling data can dominate adaptation (Figure~\ref{fig:collecct-data}). Labeling cost varies across deployment scenarios: real-world feedback incurs resource usage and user experience penalties~\cite{pensieve-sigcomm17, firm-osdi20}; simulator-based feedback is limited by simulation latency~\cite{mimicnet-sigcomm21, yang2022deepqueuenet}; and in some tasks, labels require costly human annotation~\cite{caravan-osdi24}.

\name augments existing in-network learning systems like  Caravan~\cite{caravan-osdi24}, which leverages different labeling agencies (e.g., combining LLMs and experts).  \name shifts the focus from how to acquire labels, to when and which inputs should be labeled by introducing a \emph{labeling agent}. 
Thereof, we address the tradeoff between training and data labeling costs, including when to initiate a data labeling round in Section~\ref{sec:orchestrator}, to optimize both aspects for end-to-end adaptation efficiency. 

\paragraph{Data Utility Proxy.}
Capturing the potential model improvement after labeling and training on a data sample (\ie, its data utility) is challenging due to the diversity of learning-based systems, heterogeneous input states, and the evolving nature of utility as training progresses. For example, in learned job scheduling~\cite{decima-sigcomm19}, labeling a new job may yield little benefit if the model has already seen many similar resource patterns, especially when we consider the prevalent long-tailed distributions in systems operational data, highlighting the inefficiency of naive balanced-class or coverage-based labeling strategies adopted in prior systems~\cite{caravan-osdi24}.

Given that the system's input state (e.g., job resource demands) is already available, \name uses lightweight model prediction uncertainty on the input as the utility proxy. Intuitively, the model learns most from inputs it is least confident about. 
In classification tasks (e.g., traffic classification~\cite{leo-nsdi24}), uncertainty is measured as the confidence gap between the top two predicted classes. 
In regression and RL-based tasks (e.g., flow prediction~\cite{dhukic2019-flux}, job scheduling~\cite{decima-sigcomm19}), uncertainty is quantified via the width of prediction intervals. 

\begin{figure}[t]
  \centering
  \begin{minipage}{0.48\linewidth}
    \centering
    \includegraphics[width=\linewidth]{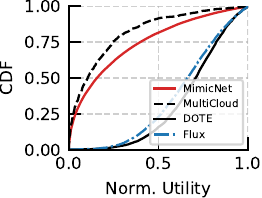}
    \caption{Data utility varies across inputs.}
    \label{fig:data-utility}
  \end{minipage}
  \hfill  
  \begin{minipage}{0.48\linewidth}
    \centering
    \includegraphics[width=\linewidth]{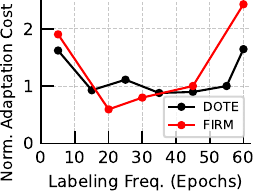}
    \caption{Too frequently or infrequently collection incurs suboptimal costs.\label{fig:freq-tradeoff}}
  \end{minipage}
\vspace{-.2cm}
\end{figure}

We validate this approach across learning-based systems, including DOTE~\cite{dote-nsdi23} (WAN traffic engineering), QCLIMB~\cite{li2024-QCLIMB} (flow prediction), and Mimicnet~\cite{mimicnet-sigcomm21} (datacenter network simulation). Figure~\ref{fig:data-utility} shows that model prediction uncertainty varies across data points, with higher uncertainty (utility) suggesting that collecting the system feedback (label) of that data input can provide more knowledge.

\paragraph{Cost-Aware Data Labeling.}
While the data utility proxy enables identifying high-utility data for labeling, naively prioritizing them can incur substantial costs due to heterogeneous labeling costs. A strawman is to formulate the problem as a budgeted knapsack: each candidate input is treated as an item, where the value corresponds to its estimated utility and the weight corresponds to its labeling cost. An integer linear programming (ILP) solver can then select the optimal subset that maximizes total utility under the budget.

However, learning-based systems introduce unique challenges. First, practical deployments must contend with \emph{outliers}. For example, system errors or noisy features may cause inputs to appear anomalously high-utility, wasting budget on mislabeled priorities.
Second, system inputs often follow \emph{long-tailed distributions} (e.g., most network flows being short~\cite{dhukic2019-flux}), so high-utility candidates may correspond to nearly identical inputs, leading to redundant labels yet limited knowledge gain.
Third, utility estimates are \emph{non-stationary}. As the model improves, the marginal benefit of labeling similar inputs evolves.

\name employs an opportunistic data-labeling strategy that samples inputs with probabilities proportional to their utility-to-cost ratio, moving beyond deterministic knapsack selection to achieve better resilience while naturally smoothing out outliers. Sampling proceeds until the round's labeling budget is consumed, ensuring robustness against noisy spikes in estimated utility. To address evolving utility, \name dynamically adjusts the per-round labeling budget in coordination with training progress, balancing the cost tradeoff between additional data collection and model refinement, which is further discussed in Adaptation Orchestrator (\S\ref{sec:orchestrator}).

\begin{algorithm}[t]
\caption{\name Adaptation Runtime}
\label{alg:orchestrator}
\DontPrintSemicolon

\KwIn{\text{unlabeled\_data}, \text{init\_model}} 
\KwOut{\text{trained\_model}, \text{updated\_labeled\_data}}
\SetKwProg{Fn}{Function}{}{}

\BlankLine
\Fn{\text{AdaptOrchestrator}():}{
  \tcp{Pre-training: reuse prior states to warm start (\S\ref{sec:transformer})}
  model, trans\_data $\gets$ \text{StateTransform}(unlabeled\_data, init\_model)\; \label{line:pretrain}

  \BlankLine
  \tcp{In-training: balance training and data labeling costs (\S\ref{sec:labeler}--\ref{sec:orchestrator})}
  \For{t in range(training\_epochs)}{ \label{line:loopstart}
    \tcp{Use proxy data to estimate collection cost-to-benefit}
      $(benefit_{label}, benefit_{cont}) \gets$ \text{EstimateBenefit}(model, trans\_data)\;

      \If{$benefit_{label} > benefit_{cont}$}{ \label{line:labeltrigger}
          (new\_labeled, budget) $\gets$ \text{LabelingAgent}(trans\_data, model, budget)\;
          labeled\_data.append(new\_labeled)\;
      }

      model $\gets$ \text{TrainStep}(model, labeled\_data)\;
  } \label{line:loopend}

  \BlankLine
  \tcp{Post-adaptation: register the learnt model and metadata of environment state for future reuse (\S\ref{sec:orchestrator})}
  \text{StateStore.Register}(trained\_model, trans\_data)\; \label{line:post}
}

\BlankLine

\Fn{\text{StateTransform}(\text{data, model}):}{
  \tcp{Align unlabeled inputs to source state space}
  source\_state $\gets$ IdentifySource(data, model)\;
  \Return{Transform(data, model, source\_state)}
}

\BlankLine
\Fn{\text{LabelingAgent}(\text{data, model, budget}):}{
  \tcp{Select samples by utility-per-cost under budget}
  data\_util $\gets$ EstDataUtility(data, model)\;
  chosen $\gets$ ProbabilisticSelect(data\_util, budget)\;
  (labeled, budget) $\gets$ Label(chosen)\;
  \Return{(labeled, budget)}
}
\end{algorithm}

\subsection{Adaptation Orchestrator: Manage States}
\label{sec:orchestrator}

The State Transformer reduces training costs by reusing system knowledge, while the Labeling Agent lowers data costs through selective acquisition. Both are essential to end-to-end efficiency (\S\ref{sec:background}), yet together they create a fundamental tension: as shown in Figure~\ref{fig:freq-tradeoff}, overly frequent collection inflates labeling costs, while infrequent collection slows convergence and wastes training cycles on diminishing-utility samples, both yielding suboptimal adaptation.

As summarized in Algorithm~\ref{alg:orchestrator}, 
\name introduces an \emph{Adaptation Orchestrator}  to optimize end-to-end adaptation efficiency. 
The \emph{State Transformer} reuses prior states to warm-start model fine-tuning (Line~\ref{line:pretrain}).  
During training, the Orchestrator continuously assesses whether to continue training or trigger new labeling based on marginal benefit-to-cost (Lines~\ref{line:loopstart}--\ref{line:loopend}), with the labeling agent enforcing cost-aware selection  (Line~\ref{line:labeltrigger}).  
Finally, after adaptation, the Orchestrator registers the adapted state into the \emph{StateStore}, managing caching, eviction, and reuse in online deployments (Line~\ref{line:post}). We next introduce how the Orchestrator coordinates training and data labeling over rounds. 

\paragraph{Balancing Training and Labeling Costs.}
As data utility evolves over training, maximizing cost-effectiveness requires progressive labeling. Yet, two coupled decisions arise: when to initiate a new labeling round, and how much budget $B_t$ to allocate at round $t$. \name resolves both decisions using a unified principle: compare the \emph{marginal utility per unit cost} of continued training (\ie, skip data labeling at $t$) with that of training on newly labeled data.  

Let $\mathcal{U}_t$ denote the total model uncertainty at round $t$, and $C_t$ the training cost. Continued training yields utility gain $\mathcal{U}_{t-1} - \mathcal{U}_t$ at cost $C_t$, \ie, total reduction in model prediction uncertainties. 
To approximate the benefit (uncertainty reduction) of labeling, \name maintains a small \emph{proxy dataset} (e.g., sampling $p$ = 1\% of the unlabeled pool). By performing data labeling on the proxy with a scaled budget $p \cdot B_t$, \name measures the uncertainty reduction on this proxy and scales it by $1/p$ to obtain an unbiased estimate of the benefit on the new labeled data: $\Delta \mathcal{U}^{\text{label}}_t=(\mathcal{U}^{\text{prox}}_{t-1} - \mathcal{U}^{\text{prox}}_t)/p$. This estimate, combined with the observed benefit on existing data, defines the projected uncertainty reduction that we could have achieved by performing one round of data collection: 
\begin{equation}
\mathcal{U}^*_t - \mathcal{U}_{t-1} = (\mathcal{U}_t - \mathcal{U}_{t-1}) + \widehat{\Delta \mathcal{U}^{\text{label}}_t}
\end{equation}
Here, the projected cost must also be adjusted as incorporating new data increases training costs.  Let $n_t$ denote the current number of trained samples and $\Delta n_t$ the number of newly added samples, estimated by the number of labeled samples in the proxy data. Then, the training cost inflates to
\begin{equation}
C^*_t = C_t \cdot \Big(1 + \frac{\Delta n_t}{n_t}\Big), \quad where \quad \Delta n_t = \frac{|\text{selected\_in\_proxy}|}{p}
\end{equation}
The orchestrator, therefore, triggers a new round only if the projected benefit-to-cost ratio of labeling surpasses that of continued training:
\begin{equation}
\frac{\mathcal{U}^*_t - \mathcal{U}_{t-1}}{C^*_t + B_t} > \frac{\mathcal{U}_t - \mathcal{U}_{t-1}}{C_t}
\end{equation}

Once triggered, the second decision is budget sizing. Instead of reusing a fixed budget, \name adapts $B_t$ online using feedback. Specifically, it tracks the realized benefit-to-cost ratio in each round,
$
\dfrac{\mathcal{U}_{t-1} - \mathcal{U}_t}{C_t + B_{t-1}},
$
and applies an additive-increase multiplicative-decrease (AIMD) update~\cite{copa-nadi18}: if the ratio drops, $B_t$ is increased to label more data to speed up model convergence, thus reducing training cost; otherwise, $B_t$ is decreased by half to avoid over-labeling. This feedback control complements the trigger condition, maximizing total adaptation cost-effectiveness with negligible overhead (\S\ref{eval:break-down}).

\paragraph{Managing State Repository in the Wild.}
Practical deployment demands continuous tracking, reuse, and evolution of environment states (e.g., due to new users or workloads). \name supports this by maintaining a lightweight \emph{state store} that caches trained models alongside a small subset of associated input data for each environment state.

\name's state management is lightweight, storing a few hundred data samples per state (\S\ref{sec:transformer}). Compared to model training, the storage cost is negligible---e.g., about \$4/TB per month on Google Cloud Platform vs.  \$11 per H100 GPU training hour, and a single state can be repurposed many times. 
To further control memory footprint, \name employs a decay-aware Least Frequently Used (LFU) eviction policy, where access frequencies decay exponentially over time, such as by 0.9 daily, favoring states that are recently and frequently reused. 
When multiple cached states are similar (small MMD distance), \name evicts the one that yields lower accuracy, preserving not only reuse but also effectiveness and resilience. Our evaluations show that \name delivers consistently good performance under tight storage constraints (\S\ref{eval:ablation}). 

Moreover, \name offers stronger compliance with data governance than existing adaptation systems~\cite{flash-mlsys24}. Whereas prior designs often require access to raw source and target data, \name identifies state mappings using only small data subsets. Once the transformation matrix $W$ is computed, clients can apply it locally without server involvement. \name further respects privacy by matching new requests to candidate states under configurable access control policies (e.g., restricting reuse to the same organization).  Moreover, state transformation operates on distributional statistics, inherently obfuscating individual samples and enabling developers to inject local noise into individual data prior to transformation as in differential privacy designs~\cite{dl-dp}. Our analytical proof in Appendix~\ref{sec:privacy} and empirical studies (\S\ref{eval:ablation}) confirm that noise addition minimally impacts transformation precision while ensuring compliance with data governance.

\section{Implementation}
\label{sec:implementation}

We have built a prototype of \name in $\sim$3,200 lines of Python. The prototype provides user-friendly APIs and a plugin interface requiring only a few lines of integration code.

\paragraph{\name Backends}
The \name backend supports execution on CPUs, with the option to leverage GPU acceleration for enhanced performance. The \emph{State Transformer}, which handles kernel-based state transformations and similarity measurements, utilizes GPUs to accelerate matrix operations. 
The \emph{Labeling Agent} is integrated into the training loop as a pluggable data source, exposing APIs for sampling, labeling, and budget allocation. It is implemented as a coroutine that can run asynchronously with training.
\name exposes model weights and dataset views through standardized PyTorch APIs, enabling systems to reuse training infrastructure with minimal changes. Coordination between the local \name agents and the central state repository uses gRPCs, ensuring scaling across machines.

\paragraph{\name Interfaces.} 
Appendix~\ref{app:apis} details the APIs through which \name complements existing learning-based systems.

\section{Evaluation}
\label{sec:eval}

We evaluate \name on eight representative state-of-the-art learning-based network systems, including job scheduling, network simulation, and adaptive bitrate streaming. Our evaluation highlights the following key findings:

\begin{denseitemize}
\item \name reduces adaptation costs (\eg, GPU compute time) by 14.9--42.4\% and improves responsiveness to runtime dynamics by 2.3--15.3$\times$, while boosting system performance (\eg, network throughput) by 6.9--31.3\% (\S\ref{eval:e2e}).

\item \name achieves effective adaptation by optimizing the sweet spot of training and data labeling efficiency (\S\ref{eval:break-down}).

\item \name augments learning-based systems over a wide range of settings and outperforms its design counterparts (\S\ref{eval:ablation}).
\end{denseitemize}

\subsection{Methodology}
\label{exp:setup}

\paragraph{Use Cases.} 
We apply \name to seven state-of-the-art learning-based networked systems spanning eight representative networking and systems applications (Table~\ref{tab:workloads}): 

\begin{denseitemize}
\item \emph{DOTE}~\cite{dote-nsdi23}: Predictive WAN traffic engineering with DL. 
\item \emph{FIRM}~\cite{firm-osdi20}: SLO-aware microservice resource management with RL agents. 
\item \emph{Flux}~\cite{dhukic2019-flux}: Flow size prediction for network scheduling using linear regression (LR), evaluated on PageRank, K-Means, and web server workloads. 
\item \emph{MimicNet}~\cite{mimicnet-sigcomm21}: Datacenter flow simulation with LSTMs on FatTree topologies. 
\item \emph{Pensieve}~\cite{pensieve-sigcomm17}: RL-based adaptive bitrate (ABR) control for video streaming. 
\item \emph{NetLLM}~\cite{netllm-sigcomm24}: LLM-based scheduling for ABR and cluster job scheduling (CJS) tasks. 
\end{denseitemize}

\begin{table}[t]
\centering
\renewcommand{\arraystretch}{1.2} 
\small  
\begin{tabular}{>{\centering\arraybackslash}m{1.7cm}|>{\centering\arraybackslash}m{4.3cm}|>{\centering\arraybackslash}m{1.4cm}} %
\hline
\textbf{Sys. Name} & \multicolumn{1}{c|}{\textbf{Task}} & \textbf{Model} \\ \hline
DOTE~\cite{dote-nsdi23} & WAN traffic engineering optimization & DNN \\ \hline
FIRM~\cite{firm-osdi20} & Resource management in microservices & RL \\ \hline
Flux~\cite{dhukic2019-flux} & Flow size prediction for scheduling & LR \\ \hline
MimicNet~\cite{mimicnet-sigcomm21} & Traffic simulation in datacenter networks & LSTM \\ \hline
IDS-LSTM~\cite{attack-data} & Streaming network intrusion detection & LSTM \\ \hline
Pensieve~\cite{pensieve-sigcomm17} & Video adaptive bitrate streaming & RL \\ \hline
NetLLM~\cite{netllm-sigcomm24} & Video adaptive bitrate streaming & LLM \\ \hline 
NetLLM~\cite{netllm-sigcomm24} & Cluster job scheduling & LLM \\ \hline
\end{tabular}
\caption{Our evaluations cover eight state-of-the-art learning-based networked systems.}
\label{tab:workloads}
\vspace{-.3cm}
\end{table}

We follow the task configurations and model choices from the original papers to ensure comparability. For example, in online Flux deployments, new jobs arrive while completed ones exit the platform. Each new job triggers an \name agent for efficient adaptation, with the model used for predicting traffic sizes during job execution, and we report its online system performance (e.g., Figure~\ref{fig:flux-online}) and average adaptation performance across jobs (e.g., Figure~\ref{fig:e2e-time}). 

We have added detailed descriptions of workload, online adaptation, and system setups in Appendix~\ref{app:setup}.

\paragraph{Evaluation Platform.} 
Because many of these systems involve resource-intensive training (e.g., NetLLM on Llama-2), we conduct experiments on a 30-node cluster designed for distributed training and online deployment. The cluster consists of 10 NVIDIA A100 GPUs (\eg, for distributed Llama-2 training) and 20 CPU nodes (\eg, for online job scheduling), each CPU node equipped with 20 cores and 192 GB DDR4 memory. We use traces (\eg, job resource demands and arrivals) from the original papers to preserve fidelity.  

\paragraph{Baselines.} 
To the best of our knowledge, \name is the first system to jointly optimize training and data efficiency for model adaptation across diverse learning-based systems and models. We compare against three baselines: 
\begin{denseitemize}

\item \emph{W/o \name}: We follow the default application setting to train their model (e.g., training from randomly initialized model weights). 

\item \emph{Caravan}~\cite{caravan-osdi24}: A state-of-the-art system for online learning in in-network systems, which selectively triggers continuous learning to maximize system performance.  

\item \emph{Flash}~\cite{flash-mlsys24}: A meta-learning-based adaptation framework for cloud platforms, requiring intrusive model modifications and limited to DL models.  
\end{denseitemize}

\paragraph{Metrics.} 
We evaluate \name along three key dimensions:  

\begin{denseitemize}
\item \emph{Adaptation Time}: The latency to adapt to new environments (\eg, model training), critical for online deployments under dynamics.  
\item \emph{Adaptation Cost}: The expense, including GPU time in training and data labeling (e.g., via simulation) costs, estimated using Google Cloud Platform (GCP) pricing.  
\item \emph{System Performance}: The final system performance (\eg, model accuracy) achieved after adaptation.  
\end{denseitemize}

All results are averaged over five runs for reliability. 

\subsection{End-to-End Performance}
\label{eval:e2e}

\begin{figure}[t]
  \centering
  \includegraphics[width=0.99\linewidth]{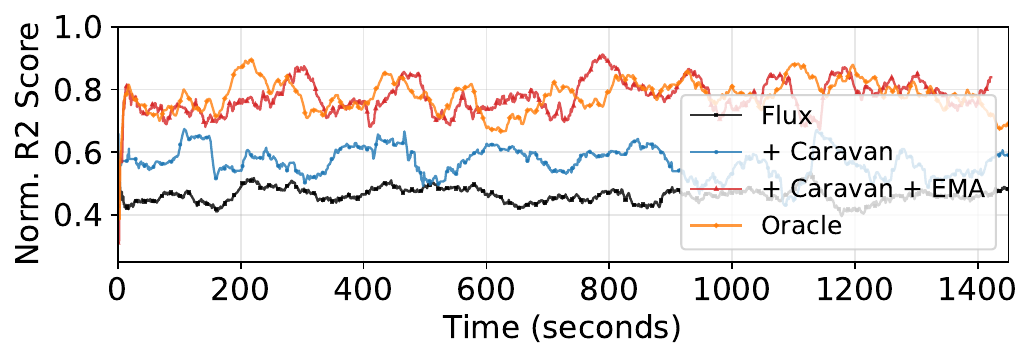}
  \vspace{-.4cm}
  \caption{In online FLUX deployments with job arrivals, \name dynamically adapts prior models from similar states, sustaining high system performance.} 
  \label{fig:flux-online}
\end{figure}

\begin{figure}[t]
  \centering
  \includegraphics[width=0.99\linewidth]{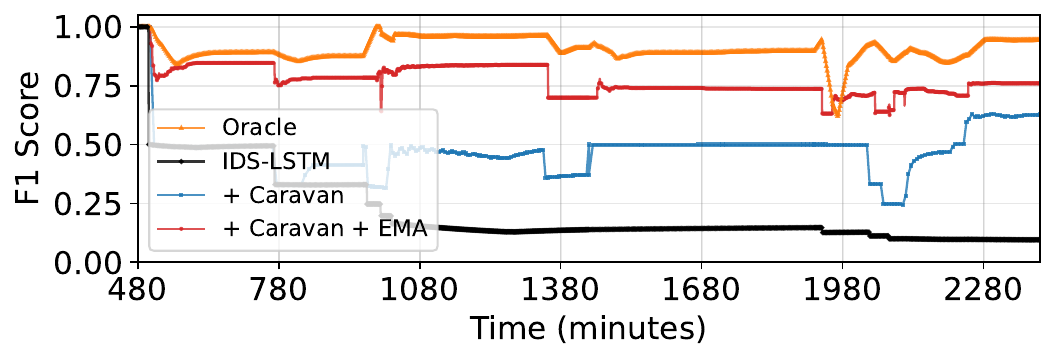}
  \vspace{-.4cm}
  \caption{In streaming environments with dynamic attack arrivals, an IDS-LSTM  adapts to distribution shifts in network traffic with \name, achieving better online intrusion detection.} 
  \label{fig:attack-online}
\end{figure}

\begin{figure}[t]
  \centering
  {
    \subfigure[Time to Accuracy on DOTE. \label{fig:T2A-dote}]{\includegraphics[width=0.48\linewidth]{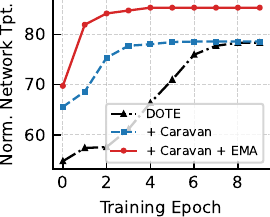}}
    \subfigure[Time to Accuracy on Flux. \label{fig:T2A-dote}]{\includegraphics[width=0.48\linewidth]{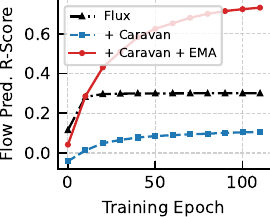}}
    \subfigure[Time to Accuracy on Firm. \label{fig:T2A-dote}]{\includegraphics[width=0.48\linewidth]{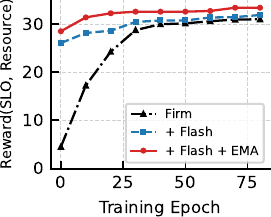}}
    \subfigure[Time to Accuracy on Pensieve. \label{fig:T2A-dote}]{\includegraphics[width=0.48\linewidth]{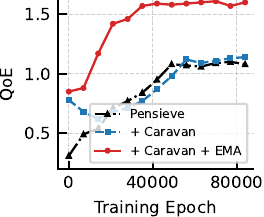}}

  }
  \vspace{-.4cm}
  \caption{\name enables faster adaptation and better post-adaptation system performance.}
  \label{fig:e2e-time}
\end{figure}

\paragraph{\name enables faster system adaptation to new environments and dynamics.} 
Figures~\ref{fig:flux-online}--\ref{fig:e2e-time} show that \name significantly accelerates adaptation across diverse learning-based systems and in online system deployments. Instead of restricting transfer to the latest global model as in Caravan~\cite{caravan-osdi24}, \name repurposes models trained on similar environment states, achieving much better starting accuracy (i.e., training epoch=0 in Figure~\ref{fig:e2e-time}). Together with informed data labeling, \name reduces the training required to reach target accuracy by 2.3--15.3$\times$, consistently outperforming both Caravan (\ie, system+Caravan) and Flash~\cite{flash-mlsys24} (\ie, system+Flash). Note that \name can augment Caravan by kicking off adaptation and selective data labeling (\ie, system+Caravan+\name). 

In online settings, \name closely tracks oracle performance (Figures~\ref{fig:flux-online} and~\ref{fig:attack-online}), where the oracle represents a hypothetical system model trained with access to future data. 
These gains generalize across a wide spectrum of model families, including LR, DL, RL, and LLMs, demonstrating \name's robustness across system designs.

\paragraph{\name improves post-adaptation system performance.} 
In addition to faster adaptation, \name improves the final system performance. As shown in Figure~\ref{fig:e2e-time}, when adaptation (\ie, model training) completes, \name raises post-adaptation system performance by 6.9--31.3\% across applications. For example, in \emph{Pensieve}, \name boosts user Quality of Experience (QoE) in ABR tasks by 31.3\%; in \emph{DOTE}, it increases network throughput by 12.5\%; and in \emph{FIRM}, it improves compound reward---a metric capturing both SLO attainment and resource utilization---by 6.9\%. These gains stem from transferring richer, environment-specific knowledge from prior deployments. The benefit becomes more pronounced under runtime dynamics: as shown in Figure~\ref{fig:flux-online} and Figure~\ref{fig:attack-online}, such as in online FLUX deployments where flow size distributions shift due to job arrivals and completions, \name repurposes prior similar environments as stronger starting points, thereby sustaining higher system performance.

\begin{figure}[t]
  \centering
  {
    \subfigure[DOTE (WAN traffic engineering). \label{fig:e2e-dote}]{\includegraphics[width=0.48\linewidth]{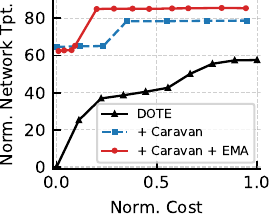}}
    \hfill
    \subfigure[Flux (Flow size prediction). \label{fig:e2e-flowpredict}]{\includegraphics[width=0.48\linewidth]{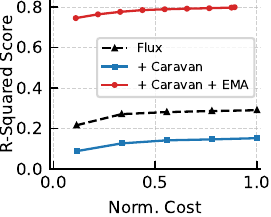}}
    \subfigure[NetLLM (Bitrate streaming).  \label{fig:e2e-netllm}]
    {\includegraphics[width=0.48\linewidth]{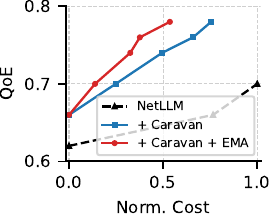}}
    \hfill
    \subfigure[FIRM (Microservices).
    \label{fig:e2e-firm}]
    {\includegraphics[width=0.48\linewidth]{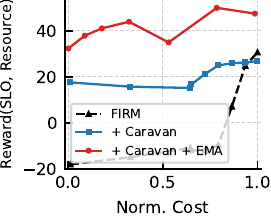}}
    }
    \vspace{-.4cm}
  \caption{\name reduces adaptation cost.}
  \label{fig:e2e-cost}
\end{figure}

\paragraph{\name reduces adaptation costs for system deployments.} Figure~\ref{fig:e2e-cost} illustrates that \name substantially lowers the total cost of model adaptation. By reducing training time and labeling fewer data, \name decreases the monetary cost of adaptation by 14.9--42.4\% relative to Caravan, based on GCP pricing models. 
While \name may require a brief warm-up phase as transformed data states stabilize, the overall adaptation remains both faster and cheaper. 
This combination of reduced training overhead and data efficiency makes \name especially attractive for large-scale and resource-constrained system deployments. Due to the space limit, we leave improvements for NetLLM's CJS task in Appendix~\ref{app:llm-cjs}. 

\subsection{Performance Breakdown}
\label{eval:break-down}

\begin{figure}
  \centering
  {
    \subfigure[Breakdown on DOTE Task. \label{fig:break-dote}]{\includegraphics[width=0.48\linewidth]{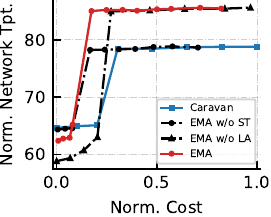}}
    \hfill
    \subfigure[Break Down on Firm Task.
    \label{fig:break-Firm}]{\includegraphics[width=0.48\linewidth]{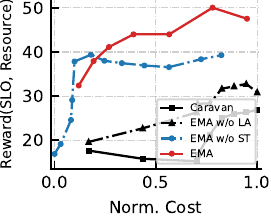}
  }
  \vspace{-.4cm}
  \caption{Performance breakdown of \name design.}
  \label{fig:breakdown-per}
 }
\end{figure}

\paragraph{Breakdown of system components.}  
To assess individual components, we evaluate two key variants of \name:  
(i) \emph{\name w/o State Transformer (ST):} This variant bypasses the State Transformer and starts tuning on the current global model.  
(ii) \emph{\name w/o Labeling Agent (LA):} This variant disables the Labeling Agent during adaptation, resorting to random data collection. Note that disabling either component automatically disables the Adaptation Orchestrator.  

Figure~\ref{fig:breakdown-per} shows that both pre-training (ST) and in-training (LA) adaptations are critical to efficiency, improving cost-to-accuracy by 10.7\% and 45.1\%, respectively. The State Transformer (\ie, \name w/o LA) effectively aligns system states, mitigating negative transfer and boosting post-adaptation system performance. Meanwhile, the Labeling Agent, together with the Adaptation Orchestrator, balances training and data collection costs, enabling more efficient adaptation.

\begin{figure}[t]
  \centering
  \begin{minipage}[b]{0.47\linewidth}
    \centering
    \includegraphics[width=\linewidth]{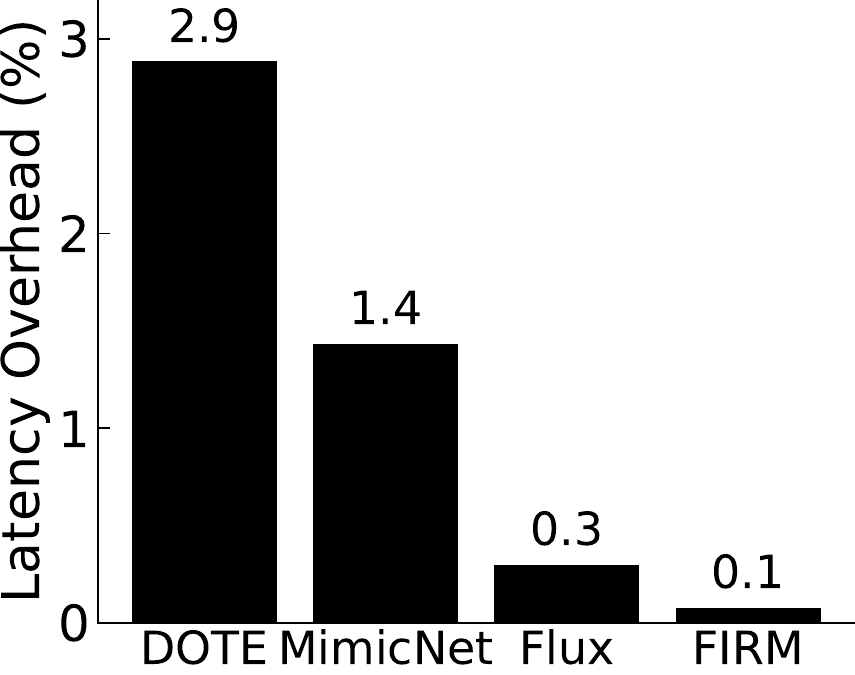}
    \caption{\name introduces little overhead.}
    \label{fig:overhead_new_1}
  \end{minipage}
  \hfill
  \begin{minipage}[b]{0.48\linewidth}
    \centering
    \includegraphics[width=\linewidth]{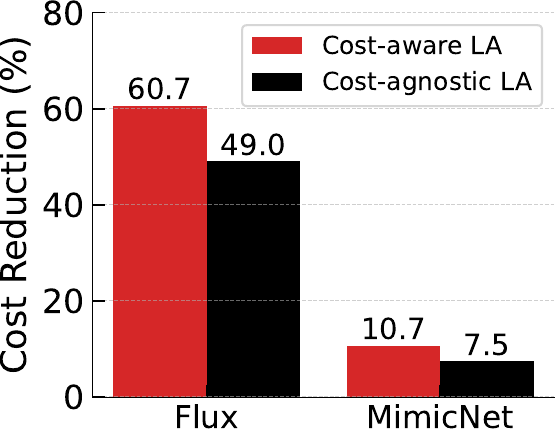}
    \caption{Accounting for heterogeneous data labeling costs is important.}
    \label{fig:cost-awareness}
  \end{minipage}
  \label{fig:cost-awareness-combined}
\end{figure}

\paragraph{\name introduces negligible system overhead.}  
As shown in Figure~\ref{fig:overhead_new_1}, \name adds only 0.3--2.9\% runtime overhead, stemming from lightweight computations in the State Transformer and Labeling Agent. This overhead is minor compared to the substantial improvements in end-to-end adaptation efficiency. Unlike prior approaches (\eg, Flash) that require intrusive modifications to model architectures, \name applies a lightweight state transformation before adaptation. Furthermore, the Labeling Agent selectively identifies high-value data while the Orchestrator minimizes collection frequency.

\begin{figure}[t]
  \centering
  \begin{minipage}[b]{0.85\linewidth}
    \centering
    \includegraphics[width=\linewidth]{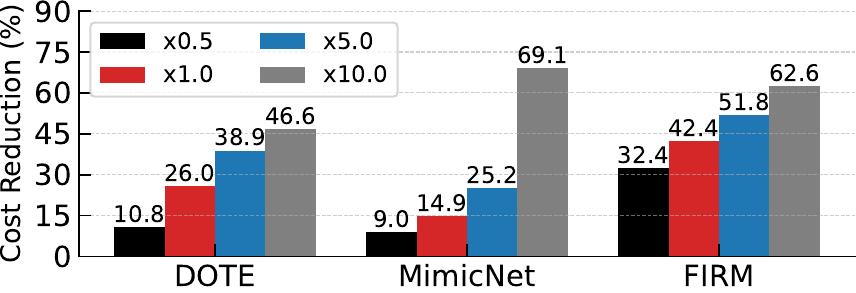}
    \caption{\name improves cost-effectiveness under different data labeling settings.}
    \label{fig:cost-factor}
  \end{minipage}
\end{figure}

\paragraph{Importance of accounting for heterogeneous labeling costs.}  
We further evaluate the impact of considering heterogeneous data labeling costs. In this experiment, we disable cost-awareness in \name's Labeling Agent (LA), selecting data solely based on utility (\ie, a cost-agnostic LA that always chooses the highest-utility samples). Figure~\ref{fig:cost-awareness} demonstrates that the cost-aware LA consistently outperforms this baseline, highlighting the inefficiency of traditional active learning approaches that ignore practical cost variations.

\subsection{Sensitivity and Ablation Studies}
\label{eval:ablation}

\paragraph{Impact of data labeling costs.}
Figure~\ref{fig:cost-factor} illustrates how \name improves the cost-effectiveness of model adaptation to reach the same post-adaptation system performance under varying data collection cost scenarios, where the labeling cost of each system input in our realistic trace is scaled by 0.5$\times$, 1$\times$, and 2$\times$. When data collection costs are low (\eg, 0.5$\times$), most savings come from reduced training costs through effective state transformation. As labeling costs increase, the benefits of \name become even more pronounced: its Labeling Agent selectively prioritizes high-utility, low-cost data, striking a balance between labeling and training costs. 

\begin{figure}[t]
  \centering  
  { 
  \subfigure[Flux. \label{fig:lw_fp}]{\includegraphics[width=0.49\linewidth]{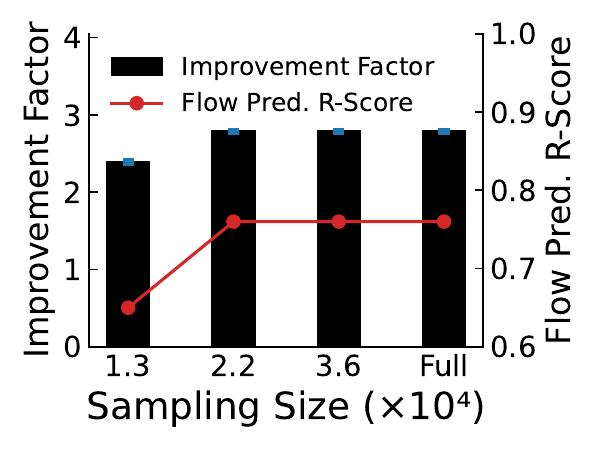}}
  \hfill
  \subfigure[DOTE. \label{fig:lw-}]{\includegraphics[width=0.49\linewidth]{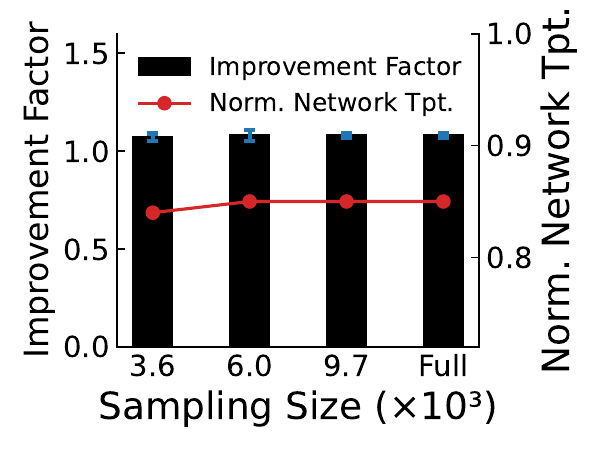}}  
  }
  \vspace{-.4cm}
  \caption{Light-weight data transformer samples data to achieve high efficiency without hurting quality.}
  \label{fig:sampling-st}
\end{figure}

\paragraph{Impact of sampling size in state transformer.}
We evaluated the effect of different sampling sizes for computing the transformation matrix $W$. As illustrated in Figure~\ref{fig:sampling-st}, increasing the number of samples improves adaptation performance, but the gains plateau beyond a few thousand samples. This observation validates our lightweight sampling approach.

\begin{figure}[t]
  \centering
  \begin{minipage}[t]{0.48\linewidth}
    \centering
    \includegraphics[width=\linewidth]{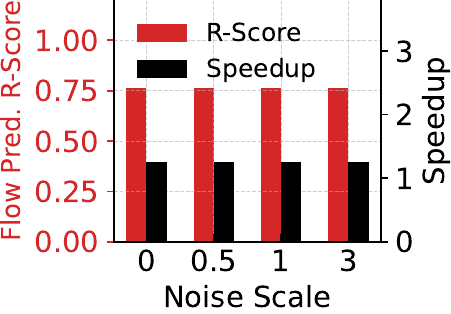}
    \caption{\name respects data privacy without hurting effectiveness.}
    \label{fig:privacy}
  \end{minipage}
  \hfill
  \begin{minipage}[t]{0.48\linewidth}
    \centering
    \includegraphics[width=\linewidth]{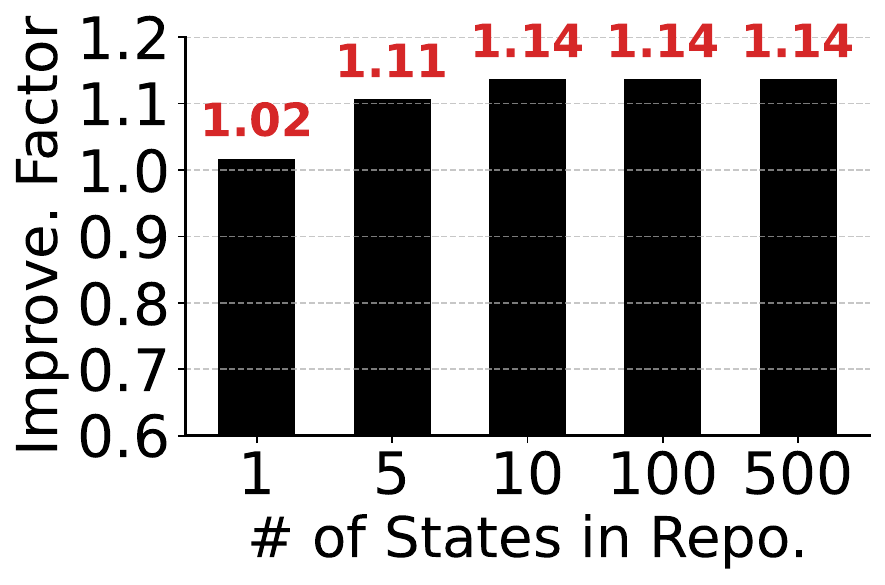}
    \caption{\name achieves improvements even with a small state repository.}
    \label{fig:repo-size}
  \end{minipage}
\end{figure}

\paragraph{Impact of privacy noise.}  
We added Gaussian noise (mean 0) to individual source data points, testing noise levels of 0, 0.5, 1, and 3 (with 0 representing no noise). As shown in Figure~\ref{fig:privacy}, \name's adaptation performance remains largely unaffected even at higher noise levels,  aligning with our theoretical analysis in Appendix~\ref{sec:privacy}, which guarantees that \name maintains effective adaptation while respecting privacy.

\paragraph{Impact of repository size.}
We evaluate \name with varying repository sizes (\ie, number of candidate parent states: 1, 5, 10, 100, 500) on the DOTE task. As shown in Figure~\ref{fig:repo-size}, \name maintains stable performance across this wide range, achieving near-optimal improvements even with a dozen parent candidates. This effectiveness stems from the state transformation module, which aligns system states without relying on highly similar states.

\section{Related Work}
\label{sec:related}

\paragraph{Machine Learning for Networked Systems.}
ML techniques have been increasingly applied to system and network optimization, typically falling into three categories.  
(i) \emph{Statistical learning methods:} QCLIMB~\cite{li2024-QCLIMB} employs random-forest-based lower-bound predictions to improve flow scheduling, while Flux~\cite{dhukic2019-flux} predicts flow sizes to enhance network management.  
(ii) \emph{Deep learning methods:} More recent efforts leverage DL for complex networking tasks, such as DOTE~\cite{dote-nsdi23} for WAN traffic optimization. Caravan~\cite{caravan-osdi24} demonstrates the potential of online learning to address non-stationary environments.  
(iii) \emph{Reinforcement learning methods:} RL has also been adopted in systems, from  Pensieve~\cite{pensieve-sigcomm17} to Astraea~\cite{liao2024astraea} for network control and adaptive video streaming.  
\name complements these systems by providing a framework for efficient adaptation under diverse,  evolving environments.

\paragraph{Domain Adaptation.}
Domain adaptation transfers model knowledge from source domains to a new target domain~\cite{long2015learning, fang2020dart}. Mutant~\cite{pappone2025mutant} leverages RL to adapt to diverse conditions by learning in real time from ongoing network interactions. 
AWARE~\cite{qiu2023aware} applies meta-learning to enable rapid adaptation to new workloads, while Flash~\cite{flash-mlsys24} embeds a meta-learner into DL architectures for cloud systems. ModelKeeper~\cite{modelkeeper-nsdi23} warm-starts model training by referring to previously trained model weights. 
However, they often require intrusive modifications, incur additional training overhead, and are primarily tailored to DL models.  Instead, \name generalizes across models and systems.

\paragraph{Selective Data Labeling.} 
Active learning (AL) aims to achieve high accuracy by selecting informative samples to annotate~\cite{settles2009active}, with methods that select samples that maximize a measure of model uncertainty, such as least confidence~\cite{lewis1995sequential} and entropy~\cite{joshi2009multi}. 
Diversity-based methods~\cite{coreset-iclr23} select a representative set of samples that span the entire feature space, using techniques such as minimum radius cover for uniform sampling~\cite{sener2017active}, clustering data to select representative points based on their distance to other samples~\cite{hacohen2022active}. Oort~\cite{oort-osdi} selects informative training data to accelerate training convergence. 
\name extends existing AL methods to learning-based systems, optimizing for scalability, cost, and end-to-end efficiency.

\section{Conclusion}
\label{sec:outtro}

This paper presents \name, a model adaptation system designed to optimize adaptation efficiency in learning-based networked systems.  \name introduces a novel state transformer to align deployment data distributions, enabling the reuse of previously trained models and substantially reducing retraining requirements.  
It further prioritizes labeling high-utility, low-cost data to minimize end-to-end adaptation costs.  
Our evaluation across eight representative learning-based network systems shows that \name reduces adaptation costs by 14.9--42.4\% while improving system performance by 6.9--31.3\%, supporting a broad range of ML models.

\label{EndOfPaper}


\bibliographystyle{ACM-Reference-Format}
\bibliography{sample-base}

\clearpage


\label{appendix}
\appendix

\section{Privacy Concern in \name}
\label{sec:privacy}

In this section, we prove that adding small noise to each data sample in the source dataset does not compromise the performance of TCA. Since TCA minimizes the Maximum Mean Discrepancy (MMD) between the source and target domains by aligning their distributions in a common feature latent space, TCA aims to make the source and target data look as similar as possible. Therefore, if adding noise to the source data does not significantly increase the MMD, it suggests that the noise has not disrupted the alignment and that TCA can still perform effectively. 

The MMD measures the difference between the source and target distributions by comparing their mean embeddings in a Reproducing Kernel Hilbert Space (RKHS). Given samples from a source domain $\mathcal{S} = \{x_i^{(s)}\}_{i=1}^{n_s}$ and a target domain $\mathcal{T} = \{x_j^{(t)}\}_{j=1}^{n_t}$, the MMD in an RKHS $\mathcal{H}$ with kernel function $k(x, y)$ is defined as:
\[
\text{MMD}(\mathcal{S}, \mathcal{T}) = \left\| \frac{1}{n_s} \sum_{i=1}^{n_s} K(x_i^{(s)}) - \frac{1}{n_t} \sum_{j=1}^{n_t} K(x_j^{(t)}) \right\|_{\mathcal{H}}
\]

\noindent where:
\begin{itemize}
    \item $K(x)$ maps $x$ into the RKHS $\mathcal{H}$,
    \item $n_s$ is the number of samples in the source domain,
    \item $n_t$ is the number of samples in the target domain.
\end{itemize}

Suppose we add zero-mean noise $\eta_i$ to each sample $x_i^{(s)}$ in the source domain. Each noisy source sample becomes:
\[
\tilde{x}_i^{(s)} = x_i^{(s)} + \eta_i
\]
where $\eta_i \sim \mathcal{N}(0, \sigma^2)$ or follows a random distribution as long as it has a mean value of 0, and each sample should be in the range such that $\|\eta_i\| \leq 1$.

The noisy source domain $\tilde{\mathcal{S}}$ can be written as:
\[
\tilde{\mathcal{S}} = \{\tilde{x}_i^{(s)} = x_i^{(s)} + \eta_i\}_{i=1}^{n_s}.
\]

\noindent We aim to show that the MMD between the noisy source domain $\tilde{\mathcal{S}}$ and the target domain $\mathcal{T}$ remains close to the original MMD between $\mathcal{S}$ and $\mathcal{T}$ by proving the following theorem:
\textbf{TCA Privacy Theorem}
\[
\text{MMD}(\tilde{\mathcal{S}}, \mathcal{T}) - \text{MMD}(\mathcal{S}, \mathcal{T}) \leq O(\|\eta_{\text{max}}\|)^2,
\]
where $\eta_{\text{max}}$ is the maximum value of $\eta_i$.

\vspace{1em}

\paragraph{Proof.}
We compute the $ \text{MMD}(\tilde{\mathcal{S}}, \mathcal{T}) $ with the noisy source data $\tilde{\mathcal{S}}$:
\[
\scalebox{0.75}{$
\begin{aligned}
&= 
\left\| 
\frac{1}{n_s} \sum_{i=1}^{n_s} \left( K(x_i^{(s)} + \eta_i) \right)
- \frac{1}{n_t} \sum_{j=1}^{n_t} K(x_j^{(t)}) 
\right\|_{\mathcal{H}}   \\
&= 
\left\|
\frac{1}{n_s} \sum_{i=1}^{n_s} 
\left( 
K(x_i^{(s)}) + J_K(x_i^{(s)}) \eta_i + O(\|\eta_i\|^2)
\right) 
- \frac{1}{n_t} \sum_{j=1}^{n_t} K(x_j^{(t)}) 
\right\|_{\mathcal{H}}  (1) \\
&= 
\left\|
\frac{1}{n_s} \sum_{i=1}^{n_s} 
K(x_i^{(s)}) - \frac{1}{n_t} \sum_{j=1}^{n_t} K(x_j^{(t)}) + \frac{1}{n_s} \sum_{i=1}^{n_s} J_K(x_i^{(s)}) \eta_i + \frac{1}{n_s} \sum_{i=1}^{n_s} O(\|\eta_i\|^2)\right\|_{\mathcal{H}} \\
&\leq
\left\| \frac{1}{n_s} \sum_{i=1}^{n_s} 
K(x_i^{(s)}) - \frac{1}{n_t} \sum_{j=1}^{n_t} K(x_j^{(t)}) \right\|_{\mathcal{H}} + 
\left\| \frac{1}{n_s} \sum_{i=1}^{n_s} J_K(x_i^{(s)}) \eta_i \right\|_{\mathcal{H}}  \\
&+ \frac{1}{n_s}\left\| \sum_{i=1}^{n_s} O(\|\eta_{\text{max}}\|^2) \right\|_{\mathcal{H}}  \\
&=  \text{MMD}(\mathcal{S}, \mathcal{T}) + 0 + O(\|\eta_{\text{max}}\|^2)
             \hspace{2.5cm} (2)
\end{aligned}
$}
\]

\noindent Hence, we have:
\[
\text{MMD}(\tilde{\mathcal{S}}, \mathcal{T}) -  \text{MMD}(\mathcal{S}, \mathcal{T}) \leq
O(\|\eta_{\text{max}}\|^2).
\]
For (1) here, we use the Taylor expansion to replace:
\[
K(x_i^{(s)} + \eta_i)
\]
with:
\[
K(x_i^{(s)}) + J_K(x_i^{(s)}) \eta_i + O(\|\eta_i\|^2),
\]
where $J_K(x_i^{(s)})$ is the Jacobian matrix of the Kernel map $K$ at point $x_i$. For (2), since $\eta_i$ is sampled from a distribution with mean value 0, we have:
\[
\frac{1}{n_s} \sum_{i=1}^{n_s} J_K(x_i^{(s)}) \eta_i = 0,
\]
and $\eta_{\text{max}}$ is also no larger than 1. 

Thus, we have shown that adding small noise to each data sample in the source dataset does not compromise the performance of TCA. Moreover, \name can maintain user privacy by adding noise to the dataset without hurting performance.

\vspace{1em}

\section{Lightweight-\name Sampling}
\label{sec:lightweight}
In this section, we give proof that sampling a subset can be lightweight without the loss of representing the full dataset distribution in order to compute the TCA transformation matrix. Specifically, we aim to determine the minimum percentage of the dataset required to ensure that the empirical cumulative distribution function (CDF) of the subset closely approximates the CDF of the full dataset within a small threshold, with high confidence. Following is a formal problem statement:

\vspace{1em}

\noindent\textbf{Problem Statement:}  
Given a dataset of size $N$ with empirical cumulative distribution function (CDF) $F_N(x)$, we aim to sample a subset of size $m$ such that the empirical CDF of the subset, $F_m(x)$, differs from the CDF of the full dataset $F_N(x)$ by no more than a small threshold $\epsilon > 0$ with high confidence $1 - \delta$. We derive the minimum subset percentage $p = \frac{m}{N}$ required to achieve this.

\begin{table*}[t]
\centering
\renewcommand{\arraystretch}{1.2}
\small
\begin{tabular}{p{3.7cm} p{13.4cm}} 
\hline
\textbf{API Name} & \textbf{Description} \\ 
\hline
\texttt{init\_service(args)} & Initialize the EMA orchestrator with configuration and resources. \\
\texttt{create\_agent(args)} & Users of the learning-based systems initiate a local client and connect to the remote \name orchestrator. \\
\texttt{transform\_state(args)} & Return dataset transformed by a pre-trained model. \textit{args} include the pretrained model and unlabeled data. \\
\texttt{label\_selection(args)} & Labeled data that is selected by Orchestrator. \textit{args} including labeled information. \\
\texttt{register(model, state)} & Register the current model and state snapshot into EMA. \\
\hline
\end{tabular}
\caption{EMA interfaces.}
\label{tab:EMA_API}
\end{table*}

\paragraph{Proof}

Let $F_N(x)$ be the empirical CDF of the full dataset:
\[
F_N(x) = \frac{1}{N} \sum_{i=1}^N \mathbf{1}\{X_i \leq x\},
\]
where $X_1, \dots, X_N$ are the samples in the dataset, and $\mathbf{1}$ is the indicator function.

Let $F_m(x)$ be the empirical CDF of a subset of size $m \leq N$:
\[
F_m(x) = \frac{1}{m} \sum_{i=1}^m \mathbf{1}\{Y_i \leq x\},
\]
where $Y_1, \dots, Y_m$ are sampled independently and uniformly from $X_1, \dots, X_N$.

Define the maximum absolute difference between the two CDFs:
\[
D = \sup_x |F_m(x) - F_N(x)|.
\]

We aim to bound $D$ such that:
\[
\mathbb{P}(D \geq \epsilon) \leq \delta,
\]
for a given threshold $\epsilon > 0$ and confidence level $1 - \delta$.

The \emph{Dvoretzky--Kiefer--Wolfowitz (DKW) inequality}
bounds the deviation between an empirical CDF and the true CDF. Applying it here, the inequality for $F_m(x)$ and $F_N(x)$ becomes:
\[
\mathbb{P}\left(D = \sup_x |F_m(x) - F_N(x)| \geq \epsilon\right) \leq 2 \exp^{-2m\epsilon^2},
\]
where $m$ is the size of the subset, and $\epsilon > 0$ is the maximum allowable difference.

Set $\epsilon = \frac{\alpha}{\sqrt{N}}$: Substituting $\epsilon$ into the inequality:
\[
\mathbb{P}\left(\sup_x \left|F_m(x) - F_N(x)\right| \geq \frac{\alpha}{\sqrt{N}}\right) \leq 2 \exp\left(-2m \cdot \frac{\alpha^2}{N}\right).
\]

 To ensure the deviation is less than $\epsilon$ with confidence $1 - \delta$, set:
\[
2 \exp\left(-\frac{2m\alpha^2}{N}\right) \leq \delta.
\]

Taking the natural logarithm of both sides:
\[
-\frac{2m\alpha^2}{N} \leq \ln\left(\frac{\delta}{2}\right).
\]

Solve for $m$:
\[
m \geq -\frac{N}{2\alpha^2} \ln\left(\frac{\delta}{2}\right).
\]

Thus, we identify a subset of size m such that sampling from this subset closely approximates the distribution of the entire dataset, enabling a lightweight computation of TCA.

\vspace{1em}

\section{Effectiveness of \name's Training-Phase Adaptation Modules }
\label{app:llm-cjs}

In this section, to evaluate the effectiveness of the EMA's training-phase adaptation mechanisms, we examine the impact of integrating the Adaptation Orchestrator and Labeling Agent into the baseline system.


\begin{figure}[t]
    \centering
    \includegraphics[width=0.5\linewidth]{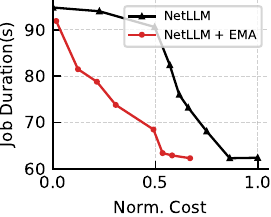}
    \caption{\name's Training-Phase Adaptation Modules improve NetLLM's performance under the same cost on the cluster job scheduling task.}
    \label{fig:cjs}
\end{figure}

\begin{figure}[t]
\begin{lstlisting}[xleftmargin=3.5ex,language=Python,label={lst:keeper_client},escapechar=|]
import EMA

def EMA_model_training(model, state, task_config):
  # Create local EMA agent based on input task config and connect to the EMA Orchestrator
  ema_agent = EMA.create_agent(task_config)

  # Transform state (input data) using EMA pre-training module
  ema_data_loader, ema_model = ema_agent.transform_state(model, state)
  
  for _ in range(num_iterations):
    # EMA guides data labeling during training
    ema_model.train(ema_data_loader, task_config)

  # Register learnt model to the Orchestrator for future reuse
  ema_agent.register(ema_model, state, task_config)
\end{lstlisting}
  \caption{EMA offers friendly APIs to enable efficient model adaptation for learning-based systems with minor changes.}
  \label{code:sys-agent}
\end{figure}

We compare EMA-augmented NetLLM against the original system on the cluster job scheduling task. Figure~\ref {fig:cjs} illustrates that equipping NetLLM with the EMA training-phase adaptation part reduces the total monetary cost by 22.5\% while achieving comparable training performance, based on execution time measured using the GCP pricing model. Moreover, at the same cost level, the EMA-augmented model consistently shows superior performance, scheduling shorter job durations. These findings highlight the essential role of adaptive data acquisition and orchestration in enabling cost-efficient adaptation in learning-based systems.

\section{\name Interfaces}
\label{app:apis}

Table~\ref{tab:EMA_API} summarizes the EMA Orchestrator API. The system is initialized with \texttt{init\_ema()}, which creates the orchestrator. \texttt{transform\_state()} processes current unlabeled data using a pre-trained model to produce a transformed unlabeled dataset. \texttt{train()} advances model training under EMA's control. When the Orchestrator determines additional data is required, \texttt{label\_selection()} labels the data with the provided ground truth. Finally, \texttt{register()} records snapshots of the current model and state for versioning and reuse. 

Figure~\ref{code:sys-agent} shows an example of \name's usage, supporting existing learning-based systems with a few lines of code change.

\section{Experiment Setup Details}
\label{app:setup}

For streaming network intrusion detection (IDS-LSTM task), we use the CIC-IDS2017 dataset~\cite{attack-data}, which contains realistic enterprise traffic collected over five consecutive days, including benign traffic and a wide range of attacks such as brute-force, DoS/DDoS, botnet activity, and port scanning. Traffic is processed in chronological order to emulate an online deployment, and the task is multi-class attack classification using an LSTM model. 
In the online evaluations, flow termination times are unknown at inference time. We therefore represent each flow using only its first 10 packets as input to the LSTM. Traffic is processed in consecutive batches, where the model is trained and evaluated per batch and continuously updated across batches to support online adaptation. 
Each packet is represented using packet-level features (e.g., payload length, TCP flags, and inter-arrival time), augmented with lightweight flow-level features available early in the flow, such as the total length of forward packets and the maximum inter-arrival time.

For NetLLM, we follow the experimental setup in the original paper~\cite{netllm-sigcomm24, decima-sigcomm19}. Specifically, the LLM is initially trained for cluster job scheduling using TPC-DS workloads, where jobs are randomly sampled from six input sizes (2, 5, 10, 20, 50, and 100\,GB) across all TPC-DS queries. This in fact leads to a heavy-tailed distribution: 23\% of the jobs contain 82\% of
the total work. We then evaluate its system adaptation to the TPC-H workload, where we sample 1,000 TPC-H jobs of six different sizes uniformly at random, and model their arrival as a Poisson process. Note that both TPC-DS and TPC-H benchmarks are popular workloads in big data systems evaluations. 

For adaptive bitrate (ABR) video streaming, we use the real-world Puffer dataset~\cite{puffer-nsdi20} collected in 2025 January. We first develop the model on a subset of user network traces, and then evaluate adaptation by deploying the learned system to different users.

\end{document}